\documentclass[11pt]{article}

\usepackage[margin=1in]{geometry}
\usepackage{amsmath,amssymb,amsfonts,mathtools}
\usepackage{booktabs,multirow}
\usepackage{graphicx}
\usepackage{enumitem}
\usepackage[protrusion=true,expansion=false]{microtype}

\usepackage{subcaption}
\usepackage[numbers,sort&compress]{natbib}
\usepackage{placeins}
\usepackage{float}
\usepackage{tikz}
\usetikzlibrary{positioning,arrows.meta,shapes.geometric,calc}
\usepackage{hyperref}
\hypersetup{colorlinks=true, linkcolor=blue, citecolor=blue, urlcolor=blue}

\title{THEIA: Learning Complete Kleene Three-Valued Logic\\
       in a Pure-Neural Modular Architecture%
       \thanks{A version of this work was accepted to the 2nd Workshop on Compositional Learning at ICML~2026 (non-archival).}}
\author{Augustus Haoyang Li\\
        Irvine Valley College\\
        Irvine, CA, USA\\
        \texttt{augustus@gus.li}}
\date{}

\begin{document}

\maketitle

\begin{abstract}
We present THEIA, a 2.75M-parameter modular neural architecture that learns the complete Kleene three-valued logic (K3) truth table from task data, without external symbolic inference components or hand-encoded K3 gate primitives. Across 5 seeds, THEIA passes all 39 entries of the full K3 diagnostic at $>99\%$ per-rule accuracy (195/195 rule--seed combinations). We do not claim that K3 learnability is unique to THEIA: appropriately trained Transformer baselines also pass all 39 rules. The central claims are instead (i) an interpretable representation profile for uncertainty propagation and (ii) a reliability spectrum for long compositional generalization under a shared discretized training protocol.

Mechanistic probing reveals uncertainty--verdict asymmetric propagation. THEIA preserves a Has-Unknown signal at every upstream engine boundary, essentially saturating the boundary-visible Bayes ceilings, while final-verdict decodability remains at or below a $73.4\%$ $U$-vs-non-$U$ oracle reference until the Logic boundary under both linear and nonlinear probes. Activation patching on matched non-absorbent configurations rules out residual-shortcut explanations for the tested Logic-Engine uncertainty-propagation cases.

A separate absorbing-state-free mod-3 sequential composition experiment generalizes from 5-step training to 500-step evaluation at $99.96\% \pm 0.04\%$ across 5 THEIA seeds. Under the identical Gumbel-softmax straight-through pipeline, flat MLPs at 0.80M and 2.75M parameters collapse to chance by 50 steps despite matching Phase~1 local accuracy; a larger-capacity ResMLP grid reaches $\geq\!99\%$ on only $3/20$ (configuration, seed) trials; and a 3.58M pre-LN Transformer reaches $99.24\% \pm 0.34\%$. The architectural contribution is therefore scoped: discretization prevents error accumulation, while the step architecture determines whether high local accuracy is sustained under end-to-end discretized training. Wall-clock convergence is reported as an auxiliary observation, not as an asymptotic optimizer-independent speed claim.
\end{abstract}

\section{Introduction}
\label{sec:intro}

Compositional reasoning---combining sub-results from independent domains into a coherent conclusion---is a hallmark of systematic generalization. Symbolic solvers such as Z3~\cite{z3} achieve formal correctness but require hand-crafted rules and cannot gracefully handle incomplete information. Neuro-symbolic systems (DeepProbLog~\cite{deepproblog}, NeurASP~\cite{neurasp}, Scallop~\cite{scallop}) bridge symbolic and neural paradigms, but in every case the actual three-valued or probabilistic inference is delegated to an external symbolic component---a probabilistic logic solver, an Answer Set Programming layer, or a Datalog engine. The neural component provides perception or differentiable scoring; the symbolic component provides the inference.

We ask two questions: \emph{can a pure neural network learn complete Kleene three-valued logic---including the non-trivial absorption rules where a definite value overrides an Unknown---without any external symbolic inference engine?} And if so, \emph{which architectural inductive biases make the resulting compositional computation reliable when evaluated on chains 100$\times$ longer than training?} These questions sit at the intersection of compositional learning and interpretability: whether modular inductive biases enable reliable compositional generalization, and whether the resulting internal representations are mechanistically transparent.

Three-valued logic with formal Unknown handling arises naturally in database query optimization (SQL NULL semantics), medical diagnosis with missing test results, and legal reasoning with undetermined facts. In all these settings, correctly propagating ``I don't know'' through a reasoning chain---and knowing when a definite value overrides that uncertainty---is essential for safe decision-making.

We present THEIA\footnote{Three-valued Hybrid Engine for Inference Architecture.}, a modular neural reasoning engine operating in 128-dimensional vector space. THEIA processes four mathematical domains---arithmetic, order relations, set membership, and propositional logic---through dedicated neural engines. Inputs may be marked as Unknown with probability $P=0.15$, and the system must propagate uncertainty according to Kleene's strong three-valued logic~\cite{kleene}---including the non-trivial short-circuit rules where a definite value absorbs an uncertain one. K3 serves here not as an application benchmark, but as a fully enumerable compositional substrate where correctness, uncertainty propagation, and internal commitment points can be verified exactly. All inference, including Unknown propagation, happens inside the network; there is no external solver.

K3 learnability is not unique to THEIA: under appropriate training, a Transformer baseline also learns the full 39-rule K3 truth table at $>99\%$ per-rule accuracy (App.~\ref{app:tuned_tf}). The central claims are instead the representation profile (uncertainty--verdict asymmetric propagation) and the reliability spectrum under discretized end-to-end composition; wall-clock convergence is auxiliary.

\paragraph{Contributions.}
\begin{enumerate}[leftmargin=*,nosep]
    \item \textbf{Uncertainty--verdict asymmetric propagation with causal localization}: The network learns to preserve the Has-Unknown signal at every upstream boundary (80.0\% / 91.1\% / 90.8\% / 99.7\% across Arith/Order/Set/Logic vs.\ $\approx\!52\%$ majority baseline), while final-verdict decodability stays at or below the $73.4\%$ $U$-vs-non-$U$ oracle reference under linear SVM and MLP probes at depths $D\!\in\!\{2,4,6\}$ (worst upstream 5-seed-mean cell $69.92\%$ at Set $d{=}6$; worst single-cell $70.09\%$ at Set $d{=}4$ seed 999; $\leq\!0.5$pp linear-to-MLP gap upstream, App.~\ref{app:mechprobe_nl}). Both signals traverse the same engines and bridges, so the asymmetry is not forced by data flow alone. Operator-identity undecodability upstream ($\approx\!20\%$ chance, App.~\ref{app:op_decomp}) is fixed by construction (Eqs.~1--4 route the operator only into the Logic Engine) and functions as a sanity check, not an independent finding. Activation patching of $\mathbf{v}_{\text{ord}}$ flips $T\!\to\!U$ on $4{,}898/4{,}898$ \emph{non-absorbent} OR pairs (5 seeds; $4{,}719/4{,}719$ on AND), ruling out residual-shortcut hypotheses for these configurations (\S\ref{sec:delayed}, App.~\ref{app:causal}).
    \item \textbf{Architectural inductive biases form a reliability spectrum for compositional generalization}: A mod-3 chain experiment with Gumbel-softmax discretization generalizes from 5-step training to 500-step evaluation at 99.96\% $\pm$ 0.04\% accuracy across 5 THEIA seeds (\S\ref{sec:chain}; per-seed range $99.90$--$99.99\%$). Controlled ablations (App.~\ref{app:backbone_ablation}) establish a four-tier reliability spectrum at 500 steps; comparator capacity ranges from $0.53\times$ (the $0.80$M flat MLP) to $2.4\times$ (the $3.58$M Transformer) of THEIA's $1.51$M chain step, with larger-capacity comparators (flat MLP $2.75$M, ResMLP $2.78$M $\pm 1.5\%$, Transformer $3.58$M) all $\geq\!1.8\times$ THEIA's capacity. Flat MLPs collapse to chance (${\sim}33\%$) despite matching Phase~1 accuracy within $0.04\%$; a ResMLP probe across a $2\!\times\!2$ depth$\times$expansion grid (4 configurations, 20 seeds) exhibits a strong depth effect (4-block mean $86$--$87\%$ vs 8-block mean $\sim\!98\%$; expansion ratio 2d vs 4d is nearly neutral at depth 8) yet reaches $\geq\!99\%$ on only $3/20$ (config, seed) trials; a pre-LN Transformer reaches $99.24\% \pm 0.34\%$ (3/5 seeds $\geq\!99\%$); THEIA reaches $99.96\% \pm 0.04\%$ (5/5 seeds $\geq\!99.9\%$; 1/5 seeds required a Phase~1 auto-restart, strict 4-seed aggregate $99.96\% \pm 0.04\%$---symmetric with the ResMLP \emph{strict} aggregation in Table~\ref{tab:resmlp_grid}). THEIA's cross-seed std ($0.04\%$) is roughly $9\times$ tighter than the Transformer's ($0.34\%$) and tens-to-hundreds times tighter than individual ResMLP configurations (per-config std $1.43\%$--$19.16\%$). The reliability gap is large under this reporting granularity, though direct comparison with lower-mean configurations is partially confounded by the bounded-metric ceiling effect (a mean near 100\% bounds the std from above; comparisons across widely differing means should be read with this caveat in mind). \emph{Task-scope note}: the 500-step result is a mod-3 cumulative-state composition task, distinct from the single-step K3 truth-table evaluation of \S\ref{sec:kleene}; Kleene chains have absorbing states ($F \wedge X{=}F$, $T \vee X{=}T$) and are deliberately not the target here (\S\ref{sec:chain} task design).
    \item \textbf{Wall-clock convergence under development-default and tuned protocols (auxiliary)}: Under per-architecture development defaults (THEIA at AdamW lr=$10^{-3}$/batch 4096; the 8L Transformer at its own default lr=$5{\times}10^{-4}$/batch 2048; Correction in App.~\ref{app:tuned_tf}), THEIA ($2.75$M parameters) reaches stable 12/12 Kleene per-rule accuracy across all 5 random seeds in $7.93 \pm 1.40$ minutes per seed. A parameter-comparable 8-layer Transformer baseline ($3.64$M params) reaches 12/12 on 7 of 8 seeds ($51.5 \pm 11.0$ minutes; the non-converging seed converges under Transformer-specific tuning, App.~\ref{app:tuned_tf}). The gap is $6.5\times$ across development defaults; a Transformer-standard tuned recipe applied to the Transformer alone ($n\!=\!3$, App.~\ref{app:tuned_tf}) narrows it to ${\sim}3.6\times$ on the Kleene-aware criterion; applying the same Transformer-standard recipe to THEIA as a partial control ($n\!=\!5$, App.~\ref{app:tuned_tf}) yields $4.93\times$ (Welch, 95\% CI $[4.40, 5.66]$). The Transformer-standard recipe slows THEIA by ${\sim}25\%$, so the third regime is a partial control, not a symmetric tuned-vs-tuned comparison; full caveat in App.~\ref{app:tuned_tf}. We list this as an auxiliary observation, not a principal contribution.
\end{enumerate}

\paragraph{Falsifiable hypotheses.} We make three concrete falsifiable claims (H1: backbone reliability under discretized composition; H2: uncertainty--verdict asymmetric propagation with causal localization; H3: optimizer-shared convergence speed), each paired with the evidence that would falsify it. Full statements and falsification criteria are in Appendix~\ref{app:hypotheses}.

\section{Related Work}
\label{sec:related}

\paragraph{Neuro-Symbolic Systems.}
DeepProbLog~\cite{deepproblog} extends probabilistic logic programming with neural predicates. NeurASP~\cite{neurasp} integrates neural networks with Answer Set Programming. Scallop~\cite{scallop} provides differentiable reasoning over Datalog. Crucially, all of these systems delegate the actual three-valued or probabilistic inference to an external symbolic component---a probabilistic logic solver, an Answer Set Programming layer, or a Datalog engine. THEIA performs all reasoning, including Unknown propagation and short-circuit absorption, neurally. Neural module networks~\cite{andreas2016} pioneered modular neural architectures for compositional reasoning, inspiring THEIA's domain-specific engine design. Marra et al.~\cite{marra2024} survey the neurosymbolic landscape. Differentiable alternatives---Logic Tensor Networks~\cite{ltn2016,ltn2022}, Neural Logic Machines~\cite{dong2019nlm}, and semantic-loss methods~\cite{xu2018semantic}---keep inference in-network but operate under two-valued or fuzzy semantics rather than K3.

\paragraph{Three-Valued Logic in AI.}
Kleene's strong three-valued logic (K3) underpins SQL's NULL handling and logic programming semantics~\cite{fitting1985}. Prior and contemporaneous work has approached K3 neural learning via architecturally-encoded gate primitives at varying granularity: the Chan--Hsu--Teh family~\cite{chan89,hsu1991twovalued,teh1995} at ordered-pair neuron activations, LNN~\cite{lnn2020} at per-neuron logical connectives with runtime symbolic inference, and independent, contemporaneous work by Damera et al.~\cite{damera2026dtlgn} at per-neuron selection over the $19{,}683$-gate $K_3$ support (see ``Differentiable logic gate networks'' paragraph below). To our knowledge, no prior system demonstrates neural learning of the complete Kleene truth table---including short-circuit absorption in both operand orders---under the stricter pure-neural criterion of \emph{generic MLP neurons with no per-neuron gate identity and no external symbolic inference component}. This framing is descriptive rather than a priority claim: we study whether K3 is learnable by an architecture meeting this criterion (random initialization, gradient descent on task samples, no hand-encoded gate semantics at the neuron level). This is distinct from in-context demonstration of K3 rules by pretrained LLMs, which inherit substantial logical priors from natural-language training data and which we do not consider to be ``learning'' in the same sense.

\paragraph{Scope of comparison.} In each reviewed line of work, prior systems either hand-encode K3 gate semantics, delegate K3 inference to an external symbolic component, or operate outside strong Kleene semantics. No K3 gate, short-circuit, or absorption rule is baked into THEIA's initialization; the 39 truth-table entries are learned from task data. The prototype classification head~\cite{snell2017} follows standard metric-learning practice and is not a K3-specific bias; the 3-class output structure is a generic classification-head choice for 3-way outputs, and all backbone comparators in App.~\ref{app:backbone_ablation} (flat MLP, ResMLP grid, Transformer) use the same 3-class prototype-matching head, so any K3-alignment benefit from the output head is shared by every architecture tested and does not favor THEIA specifically. Our ``pure neural'' framing refers to the absence of runtime symbolic delegation and of K3-specific hand-encoded primitives, not to the absence of all architectural priors. We acknowledge possible gaps in adjacent communities (multi-valued circuit synthesis, fuzzy-logic neural networks, paraconsistent-logic learning).

\paragraph{On Gumbel-softmax discretization vs.\ symbolic delegation.} One point of potential confusion: our Phase~3 sequential-composition pipeline (\S\ref{sec:chain}) uses Gumbel-softmax straight-through discretization at each chain step to snap intermediate state to one-hot K3 codes. This is a differentiable hard-argmax over the \emph{network's own output distribution}; it does not invoke an external symbolic engine and does not apply hand-encoded K3 rules at runtime. By contrast, neuro-symbolic systems such as DeepProbLog~\cite{deepproblog}, NeurASP~\cite{neurasp}, Scallop~\cite{scallop}, and LNN~\cite{lnn2020} execute symbolic inference---ProbLog resolution, answer-set grounding, Datalog evaluation, or Upward--Downward bound propagation---at runtime as a distinct algorithmic component, and in every case the K3 or probabilistic-logic semantics are specified by the user through a program or formula structure rather than learned from task gradient. Gumbel-softmax straight-through discretization invokes no rule-based inference machinery external to the forward pass; the scoping criterion of the preceding paragraph therefore accommodates discretized intermediate representations.

\paragraph{Three-Valued Neural Networks.}
Earlier work explored three-valued neural networks from a circuit-synthesis perspective. The Chan--Hsu--Teh neural-logic-network family~\cite{chan89,hsu1991twovalued,teh1995} uses an ordered pair of numbers as each neuron's activation to encode three truth values (true/false/unknown), with the three-valued connective semantics realized through architecturally-fixed combinations of these ordered-pair neurons rather than learned from data. Hsu et al.~\cite{hsu1990} generalized this to multi-valued neural logic networks with hardcoded gate semantics. The TMLNN architecture~\cite{tmlnn98} introduced training algorithms for multi-valued neuron parameters but retained hand-engineered three-valued gate primitives. More recently, Logical Neural Networks~\cite{lnn2020} hardcode each neuron as a logical connective and apply symbolic Upward--Downward inference at runtime. THEIA differs from all of these: its neurons are generic MLP units, and the Kleene three-valued algebra emerges as a learned representational property rather than as architecturally encoded primitives. LNN, the closest in design intent, cannot serve as an empirical baseline here: it requires hand-specified formula structure and does not handle numerical functions or equality, so it cannot evaluate $a+b$, $c>d$, or $c\in S$.

\paragraph{Differentiable logic gate networks.}
A parallel line of work on differentiable logic gate networks (DLGNs)~\cite{petersen2022dlgn,petersen2024convdlgn} learns compact Boolean circuits by parameterizing each neuron over the $16$ two-input Boolean gates and training a softmax-over-gates relaxation; at inference the highest-probability gate is selected per neuron. In independent and contemporaneous work, Damera et al.~\cite{damera2026dtlgn} extend DLGNs to ternary Kleene $K_3$ logic (DTLGN), where the per-neuron gate support expands to $19{,}683$ gates; they introduce Polynomial Surrogate Training (a degree-$(2,2)$ polynomial with $9$ learnable coefficients per neuron, yielding a $2{,}187\times$ parameter reduction) to make training tractable and prove a bounded gap between the polynomial surrogate and the true gate selection. DTLGN shares with THEIA the goal of neural K3 learning with an Unknown state that supports principled abstention under uncertainty, but differs in a fundamental architectural respect: DTLGN parameterizes each neuron over the K3 ternary-gate support, committing each neuron to one K3 gate post-training and therefore encoding K3 gate identity as an architectural prior at the neuron level. THEIA's neurons are generic MLP units without per-neuron gate identity; the K3 algebra emerges at the representation level (prototype-matching in a learned 128-dimensional embedding) rather than being selected from a fixed gate set. Neither approach invokes runtime symbolic delegation; the architectural axis on which they differ is whether the K3 primitive set is baked in at the neuron level (DTLGN) or left for the gradient to discover (THEIA).

\paragraph{Neural Algorithmic Reasoning and Probing.}
The CLRS benchmark~\cite{clrs} and Discrete Neural Algorithmic Reasoning (DNAR)~\cite{dnar} target deterministic two-valued algorithmic reasoning; THEIA targets three-valued logic with native Unknown propagation as a learned representational property. The length-generalization experiment of \S\ref{sec:chain} (5-step training, 500-step evaluation under Gumbel-softmax discretization) is adjacent to recent work on Transformer-based length generalization~\cite{dziri2023faith,press2022alibi,anil2022exploring} and the classical SCAN compositional benchmark~\cite{lake2018scan}, though our locality assumption (mod-3 composition without absorbing states) and mechanism (modular engines with discretized inter-step state) differ from the natural-language settings of those works. The multi-hop and modular-arithmetic results of \S\ref{sec:ablation} additionally touch on GNN expressivity~\cite{xu2020what,loukas2020} and over-smoothing~\cite{oversmoothing,oversmoothing2}. Our uncertainty--verdict propagation analysis (\S\ref{sec:delayed}) extends linear-probing methodology~\cite{alain2017} and sits within the broader mechanistic-interpretability programme~\cite{elhage2021framework,nanda2023progress} that seeks causal, circuit-level explanations of neural representations in terms of identifiable intermediate variables.

\section{Architecture}
\label{sec:arch}

\subsection{Overview}

\begin{figure*}[t]
\centering
\begin{tikzpicture}[
  scale=0.82, transform shape,
  engine/.style={
    draw, rounded corners=2pt, minimum width=1.9cm, minimum height=0.95cm,
    align=center, font=\small\bfseries, fill=blue!7
  },
  bridge/.style={
    draw, rounded corners=1.5pt, minimum width=1.3cm, minimum height=0.55cm,
    align=center, font=\scriptsize, fill=orange!15
  },
  input/.style={
    draw, circle, minimum size=0.55cm, inner sep=0pt, font=\scriptsize, fill=gray!8
  },
  ctrl/.style={
    draw, circle, minimum size=0.55cm, inner sep=0pt, font=\scriptsize, fill=yellow!20
  },
  proto/.style={
    draw, rounded corners=1pt, minimum width=0.75cm, minimum height=0.55cm,
    font=\scriptsize, fill=green!12
  },
  vec/.style={font=\small},
  arr/.style={-{Stealth[length=1.8mm]}, thick},
  darr/.style={-{Stealth[length=1.8mm]}, thick, dashed, gray!60!black}
]

\node[input] (a)   at (0, 0.55) {$a$};
\node[input] (b)   at (0, 0)    {$b$};
\node[ctrl]  (op1) at (0,-0.55) {$\oplus$};

\node[engine] (arith) at (2.1, 0) {ArithEngine};

\node[vec] (c) at (3.8, 0) {$\mathbf{c}$};

\node[bridge] (bao) at (5.7, 1.0)  {Bridge$_{ao}$};
\node[bridge] (bas) at (5.7, -1.0) {Bridge$_{as}$};

\node[input] (d)   at (7.55, 2.1) {$d$};
\node[ctrl]  (rel) at (8.45, 2.1) {$R$};
\node[input] (ss)  at (8.0, -2.1) {$S$};

\node[engine] (order) at (8.0, 1.0)  {OrderEngine};
\node[engine] (set)   at (8.0, -1.0) {SetEngine};

\node[engine] (logic) at (11.2, 0)   {LogicEngine};
\node[ctrl]   (op2)   at (11.2, 1.4) {$\odot$};

\node[engine, minimum width=1.4cm] (out) at (13.3, 0) {OutHead};

\node[proto] (pT) at (15.15, 0.6)  {$\mathbf{p}_T$};
\node[proto] (pF) at (15.15, 0)    {$\mathbf{p}_F$};
\node[proto] (pU) at (15.15, -0.6) {$\mathbf{p}_U$};

\draw[arr] (a)   -- (arith);
\draw[arr] (b)   -- (arith);
\draw[arr] (op1) -- (arith);
\draw[arr] (arith.east) -- (c.west);

\draw[arr] (c.east) |- (bao.west);
\draw[arr] (c.east) |- (bas.west);

\draw[arr] (d.south)   -- (order.north -| d);
\draw[arr] (rel.south) -- (order.north -| rel);
\draw[arr] (ss.north)  -- (set.south -| ss);

\draw[arr] (bao.east) -- (order.west);
\draw[arr] (bas.east) -- (set.west);

\draw[darr] (c.north) to[out=90, in=210] ($(order.west) + (-0.05, -0.05)$);
\draw[darr] (c.south) to[out=-90, in=150] ($(set.west) + (-0.05, 0.05)$);

\draw[arr] (order.east) -| ([xshift=-0.3cm,yshift=0.2cm]logic.west) -- ([yshift=0.2cm]logic.west);
\draw[arr] (set.east)   -| ([xshift=-0.3cm,yshift=-0.2cm]logic.west) -- ([yshift=-0.2cm]logic.west);
\draw[arr] (op2.south) -- (logic.north);

\draw[arr] (logic.east) -- (out.west);

\draw[arr] (out.east) -- (pT.west);
\draw[arr] (out.east) -- (pF.west);
\draw[arr] (out.east) -- (pU.west);

\end{tikzpicture}
\caption{THEIA architecture. Four engines process disjoint reasoning domains in a sequential--parallel topology: \textsc{ArithEngine} produces $\mathbf{c}$ from $(a, b, \oplus)$; residual bridges (orange) route $\mathbf{c}$ to \textsc{OrderEngine} (with $d, R$) and \textsc{SetEngine} (with $S$), which run in parallel and feed \textsc{LogicEngine} under operator $\odot$; \textsc{OutHead} classifies via cosine similarity against orthogonal prototypes $\{\mathbf{p}_T, \mathbf{p}_F, \mathbf{p}_U\}$. No cross-domain attention or shared parameters between engines; dashed arrows denote residual bypass.}
\label{fig:arch}
\end{figure*}
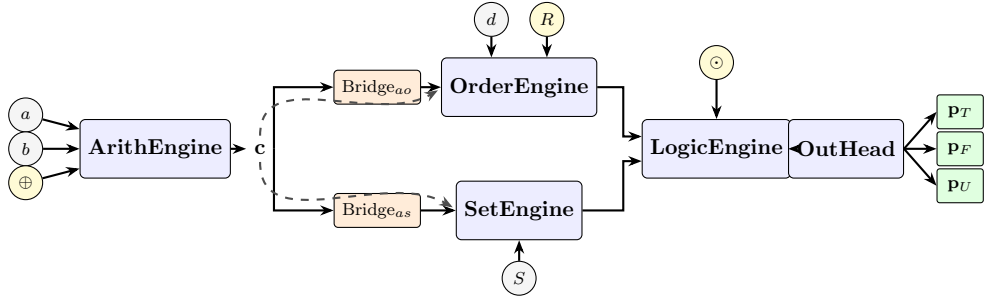

THEIA processes a four-domain reasoning chain in a fixed sequential-parallel topology (Figure~\ref{fig:arch}). We define \emph{domain-separated encoding} as an architecture where each reasoning domain (arithmetic, order, set, logic) is processed by a dedicated engine with \emph{no shared parameters} between domain encoders and \emph{no cross-domain attention}---information flows between domains only through explicit bridge layers at domain boundaries. The forward pass is:
\begin{align}
\mathbf{c} &= \text{ArithEngine}(a, b, \oplus) \\
\mathbf{v}_{\text{ord}} &= \text{OrderEngine}(\text{Bridge}_{ao}(\mathbf{c}) + \mathbf{c},\; d,\; R) \\
\mathbf{v}_{\text{set}} &= \text{SetEngine}(\text{Bridge}_{as}(\mathbf{c}) + \mathbf{c},\; S) \\
\mathbf{o} &= \text{OutHead}(\text{LogicEngine}(\mathbf{v}_{\text{ord}},\; \mathbf{v}_{\text{set}},\; \odot))
\end{align}
where $\text{Bridge}_{ao}$ and $\text{Bridge}_{as}$ are residual MLPs (Linear $\rightarrow$ GELU $\rightarrow$ LayerNorm, 33K parameters total) that transform the arithmetic output for downstream engines. The Order and Set engines operate in parallel on the bridged arithmetic vector; their outputs converge in the Logic engine. Each input has probability $P_{\text{unk}} = 0.15$ of being Unknown, receiving a learnable embedding vector. The architecture follows standard MPNN principles~\cite{gilmer2017} with domain-specific modules.

\subsection{Domain Engines}

Each engine is a small MLP stack ($\sim$0.5M parameters) with domain-specific inputs. \textbf{Arithmetic Engine}: numerical encoder fused with operator embedding. \textbf{Order Engine} and \textbf{Logic Engine}: three parallel sub-MLPs with pairwise cross-fusion, corresponding respectively to (Global/Local/Event) and (Conjunctive/Disjunctive/Implicative) subspaces. \textbf{Set Engine}: 21-dim binary vector encoder with dedicated unknown embedding. The subspace decomposition in the Order and Logic engines is an empirical design choice that improves convergence; an ablation (\S\ref{sec:ablation}) shows that replacing it with a single MLP of equivalent capacity also passes all 12 Kleene rules, indicating that the subspace structure aids convergence but is not necessary for correctness.

\subsection{Output and Multi-Hop Extension}

Three prototype vectors $\{\mathbf{p}_T, \mathbf{p}_F, \mathbf{p}_U\} \in \mathbb{R}^{128}$ serve as classification anchors in the style of prototypical networks~\cite{snell2017}; they are initialized to be mutually orthogonal (this orthogonal initialization is an engineering choice of this work, not a method from~\cite{snell2017}). Classification uses a prototype-matching head with logits $\mathbf{o} \cdot \mathbf{p}_v$, trained with cross-entropy loss and Unknown upweighting ($w_U = 2.0$). The prototypes are learned $\texttt{nn.Parameter}$s and drift from the orthonormal initialization during training---the F--T centroid cosine, for example, reaches $-0.402 \pm 0.107$ at the Logic boundary by convergence (Fig.~\ref{fig:cosine_centroids}). Consequently the dot-product logits used for training and the $\arg\max_v \cos(\mathbf{o}, \mathbf{p}_v)$ reported at inference coincide exactly only in the orthonormal limit; top-1 argmax agreement in the high-margin regime observed at convergence is expected to be tight but is not separately quantified in this work. For structural generalization experiments, a bidirectional message-passing GNN replaces the single-step Order Engine, with shared MLPs and message-passing depth $T$ decoupled from chain length.

\subsection{Domain Selection Principle}
\label{sec:domains}

The four domains---arithmetic, order, set, and propositional logic---are not an arbitrary choice but cover a minimal non-redundant set of K3-relevant atomic operations. Each domain exercises K3 Unknown propagation under a structurally distinct input-unknown-to-output-unknown mapping: numerical closure (an Unknown operand in $a+b$ produces a numerically-indeterminate result), relational comparison (an Unknown $d$ in $c > d$ produces a comparatively-indeterminate truth value), categorical membership (an Unknown $c$ in $c \in S$ produces a categorically-indeterminate membership), and propositional composition (two possibly-indeterminate operands under $\odot \in \{\wedge, \vee, \rightarrow, \leftrightarrow\}$ activate K3 absorption semantics). The four mappings are mutually non-reducible within propositional K3 over arithmetic-structured inputs: numerical indeterminacy cannot substitute for categorical indeterminacy under set membership, and so forth.

This minimality is a \emph{coverage} claim, not an \emph{exhaustiveness} claim. Extensions to quantified (first-order) logic, richer multi-valued logics (e.g., Lukasiewicz), higher predicate counts, paraconsistent systems, and DenseNet-style multi-scale residual architectures are not tested here (see \S\ref{sec:discussion} Future Work). The reliability spectrum of \S\ref{sec:chain} and App.~\ref{app:backbone_ablation}, the uncertainty--verdict asymmetric propagation of \S\ref{sec:delayed}, and the wall-clock convergence observations of \S\ref{sec:kleene} are stated within this four-domain scope. Whether they transfer to an $n$-domain generalization ($n > 4$) or to quantified extensions is left to future work.

\section{Experiments}
\label{sec:experiments}

\subsection{Setup}

All experiments use an NVIDIA RTX 5080 (16GB), AdamW optimizer~\cite{loshchilov2019} ($\text{lr}=10^{-3}$), cosine annealing, and mixed precision (FP16). Four-domain experiments use 2M samples (80/20 train-test split; 400K test samples) with $\text{NUM\_RANGE}=20$ and $P_{\text{unk}}=0.15$. The resulting label distribution is approximately 26\% False, 33\% True, and 41\% Unknown; class weighting ($w_U = 2.0$) compensates for the imbalance. Kleene diagnostic tests use 10,000 independently generated samples per rule. Diagnostic construction requires care to avoid inadvertently making operands Unknown via cross-domain coupling and to avoid edge cases in relation construction; the correct protocol and the two specific pitfalls we encountered are documented in Appendix~\ref{app:diag}. Multi-hop experiments use 1M samples with 5-hop training chains. All multi-seed experiments use seeds $\{42, 123, 256, 777, 999\}$; Transformer baseline experiments are extended to 8 seeds with $\{31415, 27182, 14142\}$ where noted.

\paragraph{Transformer baseline configurations.} We use two 8-layer, 8-head, $d{=}192$ Transformer variants~\cite{vaswani2017} (3.58--3.64M parameters): a post-LN configuration (BigTransformer) for the Kleene-task and probing comparisons (\S\ref{sec:kleene}, \S\ref{sec:delayed}), and a pre-LN variant~\cite{xiong2020preln} (TF8LTuned) for the chain pipeline (\S\ref{sec:chain}); the post-LN BigTransformer fails to converge under the chain pipeline's shared learning rate in the three-phase Gumbel-softmax setting. Details in App.~\ref{app:backbone_ablation}. \emph{Plateau-restart symmetry:} for the \S\ref{sec:kleene} development-default 4-domain comparison, plateau-restart is disabled for both THEIA and the Transformer baseline (neither training script contains a restart mechanism); the seed-123 Transformer non-convergence under the development-default setting therefore reflects 150-epoch budget exhaustion under the no-restart regime applied symmetrically to both architectures, not an asymmetric disclosure. For the \S\ref{sec:chain} chain pipeline, plateau-restart is enabled for both THEIA and TF8LTuned (via a shared Phase~1 implementation); this asymmetry between the 4-domain and chain settings is a property of the training pipeline, not of the architecture comparison within each setting.

\subsection{End-to-End Three-Valued Algebraic Learning}
\label{sec:kleene}

\begin{table*}[t]
\centering
\caption{Four-domain reasoning accuracy, each architecture at its development default. THEIA statistics are over 5 seeds; Transformer statistics are over 8 seeds (the initial 5 plus 3 additional seeds $\{31415, 27182, 14142\}$). Both architectures pass all 12 Kleene diagnostic rules at $>99\%$ on every converged seed (Table~\ref{tab:kleene}); the Transformer converges on 7/8 seeds (the one failure, seed 123, converges under Transformer-specific tuning; Appendix~\ref{app:tuned_tf}). A tuned follow-up ($n\!=\!3$, Appendix~\ref{app:tuned_tf}) narrows the $6.5\times$ wall-clock ratio to ${\sim}3.6\times$ on the Kleene-aware criterion; all three tuned seeds also pass 12/12 Kleene rules. \emph{Each architecture ran its own development default (THEIA: AdamW $\text{lr}=10^{-3}$, cosine, batch 4096; the 8L Transformer: AdamW $\text{lr}=5{\times}10^{-4}$, cosine, batch 2048), not a single shared optimizer configuration; this is a development-defaults comparison (Correction in Appendix~\ref{app:tuned_tf}), and the controlled same-recipe anchor is the recipe-applied-to-both setting. The tuned follow-up (${\sim}3.6\times$) and the recipe-applied-to-both control ($4.93\times$, Appendix~\ref{app:tuned_tf}) report the same comparison under Transformer-preferred and symmetric-recipe settings. Seed 42's wall-clock uses a 2026-04-15 doubly-fixed-diagnostic retrain; checkpoint provenance for seed~42 and its scope across analyses are documented in Appendix~\ref{app:diag}.}}
\label{tab:accuracy}
\begin{tabular}{lcccc}
\toprule
Model & Accuracy & Parameters & Seeds 12/12 & Time (min) \\
\midrule
THEIA (modular)        & 99.96\% $\pm$ 0.01\% & 2.75M & 5/5 & $\mathbf{7.93 \pm 1.40}$ \\
Transformer (8L, 8H)   & 99.98\% $\pm$ 0.01\% & 3.64M & 7/8 & $51.5 \pm 11.0$\,\textsuperscript{$\dagger$} \\
\midrule
\multicolumn{4}{r}{\textit{THEIA speedup, development defaults (tuned: ${\sim}3.6\times$, App.~\ref{app:tuned_tf})}} & $\mathbf{6.5\times}$ \\
\bottomrule
\end{tabular}

\vspace{0.5em}
\footnotesize
$^\dagger$Transformer: 7 of 8 seeds converged within 34--63 min (mean 51.5 $\pm$ 11.0 over the 7 converged seeds; the 8 attempted seeds are $\{42, 123, 256, 777, 999, 31415, 27182, 14142\}$ with seed 123 failing to converge under the development-default setting; 150-epoch budget). Seed 123 failed to converge under the development-default setting but converges under Transformer-standard tuning (Appendix~\ref{app:tuned_tf}), suggesting hyperparameter sensitivity rather than an architectural limitation.
\end{table*}

Under a shared early-stopping criterion (overall accuracy $>99.9\%$ AND all 12 Kleene rules $>99\%$ on two consecutive checkpoints), the 8L Transformer reaches 12/12 on 7 of 8 seeds with mean wall-clock $51.5 \pm 11.0$ min (range 34--63 min, over the 8 attempted seeds $\{42, 123, 256, 777, 999, 31415, 27182, 14142\}$). On the remaining seed (123) it fails to converge within the 150-epoch budget; this seed converges under Transformer-specific tuning (Appendix~\ref{app:tuned_tf}), indicating hyperparameter sensitivity rather than an architectural limitation. THEIA reaches 12/12 on 5/5 seeds in $7.93 \pm 1.40$ min, yielding a $6.5\times$ speedup across development defaults. We use a Transformer baseline of comparable rather than strictly matched capacity (3.64M vs THEIA's 2.75M, ${\sim}32\%$ parameter advantage); THEIA's efficiency is therefore not explained by capacity, although a per-parameter normalization yields a somewhat smaller apparent gap: THEIA at $2.88$ min/M-param vs Transformer at $14.1$ min/M-param $\approx\!4.9\times$. We report the unnormalized $6.5\times$ as the headline figure since total wall-clock matches the practitioner-facing question (``how long before the model is ready?''), but flag the $4.9\times$ per-parameter figure for readers who prefer a capacity-controlled metric.

\paragraph{Why Kleene accuracy is the primary metric.} Kleene short-circuit and absorption rules occur at low natural frequency ($F\!\vee\!F < 0.8\%$ of training samples), so bulk overall accuracy can be satisfied while edge-case rules remain unreliable. We therefore use the per-rule Kleene diagnostic as the primary metric and a Kleene-aware stopping criterion throughout.

\paragraph{Robustness to the stopping criterion.} The Kleene-aware criterion (overall $>\!99.9\%$ \emph{and} all 12 Kleene rules $>\!99\%$ on two consecutive checkpoints) was read off THEIA's convergence trajectory, raising a natural concern that it may favor THEIA by construction. We therefore re-measure the development-defaults gap under a Kleene-independent criterion: first-epoch wall time to overall validation accuracy $\geq\!99.9\%$ (threshold only, no Kleene requirement). On the initial 5-seed set $\{42, 123, 256, 777, 999\}$, THEIA reaches this milestone in $5.7 \pm 1.4$ min ($n{=}5$) and the development-default Transformer in $39.5 \pm 7.1$ min ($n{=}4$; seed 123 excluded as it fails to converge under the development-default setting), a ratio of $7.0\times$---wider than the $6.5\times$ Kleene-aware headline, not narrower. At looser thresholds the ratio widens monotonically ($9.8\times$ at $\geq\!99.5\%$; $12.2\times$ at $\geq\!99.0\%$). The headline gap is therefore not an artifact of the stopping criterion's alignment with THEIA's trajectory. This Kleene-independent re-analysis uses the original 5-seed set; it has not been extended to the full 8-seed Transformer baseline of Table~\ref{tab:accuracy}.

\paragraph{Hyperparameter sensitivity of the convergence gap.} A natural concern is that both architectures trained under a single optimizer configuration (AdamW lr=$10^{-3}$ + cosine) to isolate the architectural variable. To bound Transformer-specific tuning, we rerun the 8L Transformer baseline with a Transformer-standard recipe (App.~\ref{app:tuned_tf}). Under the Kleene-aware milestone, the tuned Transformer converges at ${\sim}$3.6$\times$ THEIA's (default) wall-clock (under the overall-accuracy milestone, ${\sim}$7.0$\times$; see App.~\ref{app:tuned_tf} for why the two diverge). The development-default seed 123 that failed under its development-default optimizer converges readily under tuning, confirming optimizer sensitivity rather than architectural limitation. As a partial further control, we additionally apply the same Transformer-standard recipe to THEIA (App.~\ref{app:tuned_tf} ``Transformer-recipe-applied-to-both control''), yielding a ratio of $4.93\times$ (5 seeds, Welch, 95\% CI $[4.40, 5.66]$), positioning between the development-defaults $6.5\times$ and the Transformer-only-tuned $3.6\times$. We note this is not a genuinely symmetric tuned-vs-tuned comparison: the Transformer-standard recipe \emph{slows} THEIA (from $7.93$ to $9.88$ min, a ${\sim}25\%$ degradation), so THEIA is run at a suboptimal configuration; we did not perform a THEIA-specific sweep, and a true symmetric comparison is left to future work.

\paragraph{Targeted diagnostic verification.} We construct samples with precisely controlled inputs to verify each Kleene rule independently (10K samples/rule; a rule \emph{passes} if per-rule accuracy exceeds 99\%). All 12 Unknown-involving rules pass at $>99\%$ across all 5 seeds (60/60 rule--seed combinations); commuted absorption rules ($U\!\wedge\!F^{\dagger}$, $U\!\vee\!T^{\dagger}$) are verified correctly in both operand orders (means 99.93\%), supporting the short-circuit claim under operand commutation. Extending the diagnostic to the \textbf{full 39-rule Kleene K3 truth table} (App.~\ref{app:kleene_full}) yields \textbf{195/195 rule--seed combinations passing at $>99\%$} (grand mean 99.88\%), supporting the ``complete Kleene'' claim at the full truth-table level. The tuned BigTransformer baseline likewise passes all 39 rules at $>99\%$ across three seeds (117/117, App.~\ref{app:tuned_tf}), providing a matched full-coverage comparison.

\subsection{Uncertainty--Verdict Asymmetric Propagation: Probe Evidence and Causal Verification}
\label{sec:delayed}

We extract hidden representations at \emph{domain boundaries}---the output of each engine after its full internal computation---and analyze them using linear SVM probes and inter-class Euclidean distance (Table~\ref{tab:delayed}). All probing analyses use the same 5 checkpoints as Table~\ref{tab:kleene}, 50K samples per checkpoint; results are mean $\pm$ std across seeds. Since the logic operator enters only at the Logic Engine by construction (Eqs.~1--4), upstream probes for the operator or the operator-conditional verdict are partially architecturally bounded; the Has-Unknown signal, by contrast, can reach all upstream boundaries through the input flags on $a, b, d$. The non-trivial finding is the combination of a bounded verdict side with a high-decodability uncertainty side along the same pathway.

\begin{table*}[t]
\centering
\caption{Verdict decodability and representational separation by domain boundary. Mean $\pm$ std over 5 trained checkpoints (seeds $\{42, 123, 256, 777, 999\}$); 50K samples per checkpoint. SVM accuracy is the linear probe accuracy on a 3-class classification task; F--T distance is the Euclidean distance between class centroids in the 128-dim representation space; separation ratio is the mean of per-seed (Logic/Arithmetic) F--T distance ratios.}
\label{tab:delayed}
\begin{tabular}{lccc}
\toprule
Domain Boundary & SVM Accuracy & F--T Distance & Separation Ratio \\
\midrule
Arithmetic & 0.609 $\pm$ 0.001 & 0.146 $\pm$ 0.015 & --- \\
Order      & 0.672 $\pm$ 0.002 & 3.130 $\pm$ 0.302 & --- \\
Set        & 0.697 $\pm$ 0.002 & 5.967 $\pm$ 0.643 & --- \\
\textbf{Logic} & $\mathbf{0.999 \pm 0.000}$ & $\mathbf{272.738 \pm 37.070}$ & $\mathbf{1{,}898\times}$ \\
\bottomrule
\end{tabular}
\end{table*}

\begin{figure}[!ht]
\centering
\includegraphics[width=\textwidth]{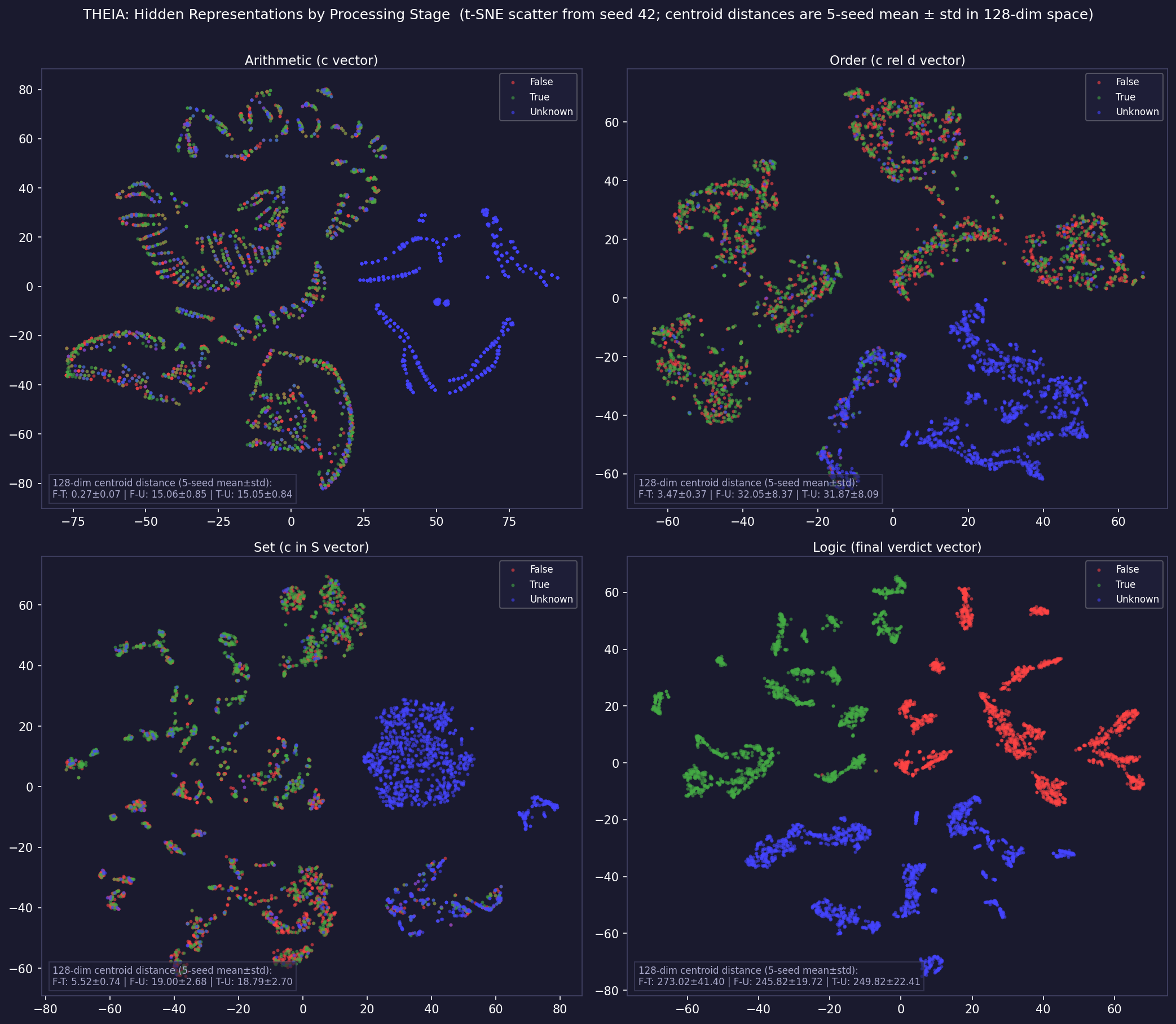}
\caption{t-SNE projection of hidden representations at the four domain boundaries (Arithmetic, Order, Set, Logic), colored by final verdict label. Points pooled across 5 canonical checkpoints (seeds $\{42, 123, 256, 777, 999\}$); $3{,}000$ random test samples per seed per boundary. t-SNE run with perplexity $=30$, random seed $=42$, $1000$ iterations. \emph{Visual pattern}: the Unknown class (blue) forms a visually separated region at every upstream boundary (Arith/Order/Set), while True (green) and False (red) are visually inseparable until the Logic boundary, where all three classes form distinct clusters. This qualitatively corroborates the Has-Unknown vs.\ final-verdict probe asymmetry of Table~\ref{tab:delayed} and Appendix~\ref{app:mechprobe}. \emph{Caveat}: t-SNE preserves local rather than global structure, so inter-cluster distances and absolute positions should not be read quantitatively; the load-bearing qualitative claim (Unknown-only separation upstream vs.\ three-way separation at Logic) is further corroborated by the centroid-direction analysis of Figure~\ref{fig:cosine_centroids}, which does not depend on a non-linear embedding.}
\label{fig:tsne_domains}
\end{figure}

\begin{figure}[!ht]
\centering
\includegraphics[width=\textwidth]{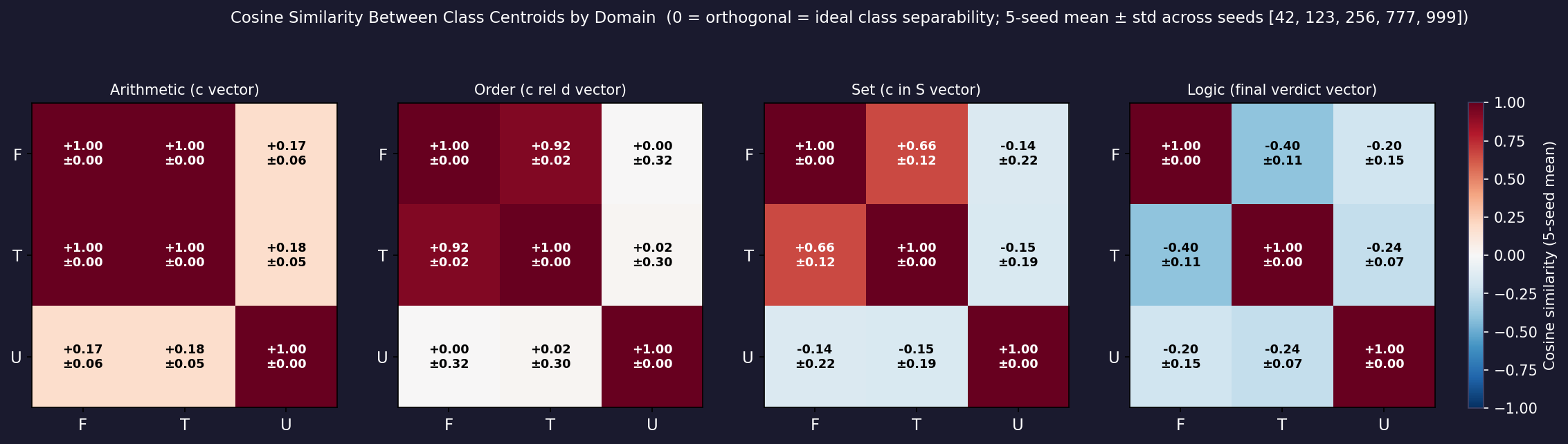}
\caption{Cosine similarity between class centroids (False, True, Unknown) at each domain boundary. Mean $\pm$ std over 5 canonical checkpoints (seeds $\{42, 123, 256, 777, 999\}$), 50K samples per seed per boundary. Off-diagonal cosine $\approx 0$ indicates orthogonal centroid directions; negative values indicate centroids pointing apart. Four observations: (i) At the Arithmetic boundary, F--T cosine $= +0.999 \pm 0.001$: the False and True centroid directions are nearly colinear, consistent with the final verdict being a function of downstream operands ($d, R, S, \odot$) not yet available at this stage, so the Arithmetic Engine has no task-driven gradient to separate F from T at the representation level. This is a stronger statement than the corresponding linear-probe accuracy of $60.9\%$ (near the non-Unknown majority): it rules out centroid-direction separation, not merely one particular probe family. (ii) F--U and T--U at Arithmetic are $+0.175 \pm 0.055$ and $+0.178 \pm 0.053$ respectively, indicating partial Unknown encoding even before the Logic Engine, consistent with the $80.0\%$ Has-Unknown probe accuracy (Table~\ref{tab:delayed}). (iii) F--T similarity decreases monotonically through the pipeline ($+0.999 \to +0.923 \to +0.663 \to -0.402$), with the sharpest drop at the Logic boundary; standard deviations broaden from $0.001$ (Arith) to $0.124$ (Set) and $0.107$ (Logic), reflecting seed-level variability in the intermediate Set/Logic representations even as the directional pattern is consistent. (iv) At Logic, all off-diagonal cosines are negative (F--T $= -0.402 \pm 0.107$, F--U $= -0.198 \pm 0.145$, T--U $= -0.240 \pm 0.070$), i.e., centroids point apart beyond orthogonality; this is the expected effect of cross-entropy training pushing class centroids past the orthogonal-prototype initialization to widen the softmax margin. The centroid-direction pattern documented here is invariant to the Euclidean-rescaling caveats discussed for the $1{,}898\times$ F--T distance ratio above.}
\label{fig:cosine_centroids}
\end{figure}

\paragraph{Verdict-side vs.\ uncertainty-side decodability.} At the Arithmetic boundary, linear-probe accuracy for the 3-class verdict is 60.9\%, below the $73.4\%$ accuracy achievable by an oracle that perfectly separates Unknown from non-Unknown samples (labeling every $VU{=}U$ correctly and predicting the majority class on the rest). This \emph{$U$-vs-non-$U$ oracle reference} is $0.4027 \times 1.0 + 0.5973 \times 0.5550 = 0.734$, where $0.4027$ is the empirical fraction of $VU{=}U$ samples and $0.5550$ is the True-class fraction among $VU{\neq}U$ samples (joint distribution measured at $N{=}50{,}000$, data seed $999$; see App.~\ref{app:mechprobe_nl}); it bounds any classifier whose verdict information is exhausted by the $U$/non-$U$ binary split, and serves here as an operational benchmark. Note that this reference is not equivalent to a predictor that sees only the input-level Has-Unknown flag: Kleene absorption can make the verdict definite even when inputs carry Unknown, so the $U$-vs-non-$U$ label is a strictly richer signal than the input-side Has-Unknown indicator. The Order and Set boundaries reach 67.2\% and 69.7\%, still below this reference; accuracy then jumps to 99.9\% at the Logic boundary. A natural alternative explanation is that verdict information is present but linearly inaccessible; to rule this out, we train multi-hidden-layer MLP probes at depths $D\!\in\!\{2,4,6\}$ ($128\!\to\!256\!\to\!\ldots\!\to\!256\!\to\!3$, GELU, dropout 0.1) on the same representations and 5 seeds, under a stricter random 60/20/20 train/val/test split with val-best-epoch selection (test evaluated once; App.~\ref{app:mechprobe_nl}). The nonlinear probe reaches 61.2\% / 67.2\% / 69.8\% / 99.9\% at Arith/Order/Set/Logic at $d\!=\!2$ (and stays below 70.0\% upstream across $d\!\in\!\{2,4,6\}$; worst upstream 5-seed-mean cell is Set at $d\!=\!6$, $69.92 \pm 0.12\%$; worst single-cell across the full seed$\,\times\,$depth grid is $70.09\%$ at Set, $d\!=\!4$, seed 999)---a worst-case gap of $+0.5$pp over the linear probe and never crossing the $73.4\%$ reference upstream (App.~\ref{app:mechprobe}, Table~\ref{tab:mechprobe_nl}). By contrast, the Has-Unknown signal is decodable at every upstream boundary well above the $\approx\!52\%$ majority baseline ($P(HU{=}0){=}0.5254$): $80.0\% / 91.1\% / 90.8\% / 99.7\%$ at Arith / Order / Set / Logic (std $<\!0.001$). Figure~\ref{fig:tsne_domains} visualizes this asymmetry directly: the Unknown class forms a visually separated region at every upstream boundary, while True and False remain visually inseparable until the Logic boundary.

\paragraph{Theoretical ceiling on Has-Unknown decodability.} To make the uncertainty-side claim precise, we compute the layer-visible information-theoretic (Bayes) ceiling on Has-Unknown probe accuracy under the 4-flag Unknown generator of this work (atomic flags $a_u, b_u, d_u, s_u$ each i.i.d.\ $\mathrm{Bernoulli}(0.15)$; empirical marginals within $0.003$ of $0.15$, App.~\ref{app:mechprobe_nl}). The optimal binary predictor at each boundary marginalizes over flags not yet injected at that boundary. Closed-form ceilings (derivations in App.~\ref{app:hu_ceilings}, Table~\ref{tab:hu_ceiling_compare}): Arith (visible $(a_u, b_u)$, $k{=}2$) gives $P(a_u \lor b_u) + P(\neg(a_u \lor b_u)) \cdot \max(P(d_u \lor s_u), P(\neg(d_u \lor s_u))) = (1-0.85^2) + 0.85^2 \cdot 0.85^2 = 0.7995$; Order (visible $(a_u, b_u, d_u)$, $k{=}3$) gives $(1-0.85^3) + 0.85^3 \cdot 0.85 = 0.9079$; Set (visible $(a_u, b_u, s_u)$, $k{=}3$) gives $0.9079$ identically; Logic (visible all four flags, $k{=}4$) gives $1.0$. Empirical per-boundary probe accuracies are $80.0\%$ / $91.1\%$ / $90.8\%$ / $99.7\%$ (clean-split val-best-ep, 5-seed mean; cross-seed std $<\!0.01\%$ upstream, reflecting that upstream hidden states are deterministic functions of the input at finite sample). Gaps to the empirical Bayes ceiling (Table~\ref{tab:hu_ceiling_compare}) are: Arith $+0.05$pp, Set $+0.02$pp (both at the ceiling within noise), Order $+0.25$pp (a small but consistent positive gap, discussed below), Logic $-0.24$pp (probe undershoots the $100\%$ ceiling, consistent with MLP approximation of an exact 4-bit lookup). \emph{Reading}: at Arith and Set the probe saturates the information-theoretic ceiling; at Logic it is near-saturating; at Order the probe very slightly exceeds the closed-form ceiling, which we attribute to one of (i) probe-selection noise under val-best-epoch, (ii) minor gradient-driven leakage of $s_u$-correlated features into $\mathbf{v}_{\text{ord}}$ via the end-to-end verdict loss, or (iii) the probe exploiting joint structure between visible flags and arithmetic values beyond the pure OR rule. All three explanations preserve the verdict-side asymmetry claim; the Order excess is $<\!0.3$pp under both the paper-original 70/30 and the clean-split val-best protocols and does not affect the load-bearing contrast that Has-Unknown decodability remains strongly above the $\approx 52\%$ majority baseline at every upstream boundary while verdict decodability stays strictly below the $U$-vs-non-$U$ oracle reference.

The unknown flags on $a, b, d$ traverse the same engines and bridges that the final verdict would traverse; the verdict-side and uncertainty-side asymmetry therefore is not a consequence of routing structure alone. The network has learned to preserve input-level uncertainty through each engine (albeit incompletely, per the theoretical ceiling above) while deferring verdict commitment to the Logic boundary---the representational signature of THEIA's compositional strategy under K3 semantics.

The F--T centroid distance grows by a factor of $1{,}898\times$ from arithmetic to logic (mean of per-seed ratios; 95\% non-parametric bootstrap CI $[1626, 2255]$, 1000 resamples), consistent with each upstream engine computing its designated variable in a representation that does not yet carry the final verdict. This ratio is a within-THEIA contraction-to-expansion pattern between engine-output representations; absolute Euclidean scale depends on intervening LayerNorm placement, and raw Euclidean distances in Table~\ref{tab:delayed} are not comparable across architectures (THEIA hidden 128, Transformer 192). The primary load-bearing claim is the linear and nonlinear probe accuracy pattern ($\leq\!69.9\%$ upstream 5-seed-mean, $\leq\!70.09\%$ worst single cell; $99.9$--$100.0\%$ at Logic), which is invariant to Euclidean rescaling. The F--T cosine similarity trajectory ($+0.999 \to +0.923 \to +0.663 \to -0.402$ across Arith/Order/Set/Logic; 5-seed mean, Figure~\ref{fig:cosine_centroids}) provides a rescaling-invariant view of the same pattern.

\paragraph{Transformer baseline layer-wise probing.} For direct comparison, we apply the same probing protocol layer-by-layer to the (larger) 8-layer Transformer baseline of Table~\ref{tab:accuracy}, which reaches the same final correctness as THEIA on the Kleene diagnostic. The probing analysis uses the development-default Transformer configuration of \S\ref{sec:kleene}; whether the contraction--expansion trajectory described below persists under the tuned configuration of Appendix~\ref{app:tuned_tf} is left to future work. The trajectory (Table~\ref{tab:tf8l_probing}) differs qualitatively from THEIA's; we discuss the implications in \S\ref{sec:discussion}.

\begin{table}[!ht]
\centering
\small
\setlength{\tabcolsep}{4pt}
\caption{Layer-wise linear probing of the 8-layer Transformer baseline (Table~\ref{tab:accuracy}), analogous to Table~\ref{tab:delayed}. Mean $\pm$ std over 5 seeds; 50K samples per seed. F--T distances are Euclidean; post-LN normalization may contribute to the layer-1 contraction.}
\label{tab:tf8l_probing}
\begin{tabular}{@{}lcc@{}}
\toprule
\textbf{Layer} & \textbf{SVM Accuracy} & \textbf{F--T Distance} \\
\midrule
input embed. & 0.750 $\pm$ 0.001 & 0.381 $\pm$ 0.011 \\
layer 1      & 0.781 $\pm$ 0.007 & 0.224 $\pm$ 0.025 \\
layer 2      & 0.785 $\pm$ 0.011 & 0.342 $\pm$ 0.097 \\
layer 3      & 0.859 $\pm$ 0.040 & 1.305 $\pm$ 0.514 \\
layer 4      & 0.999 $\pm$ 0.000 & 8.322 $\pm$ 3.113 \\
layer 5      & 1.000 $\pm$ 0.000 & 13.137 $\pm$ 0.479 \\
layer 6      & 1.000 $\pm$ 0.000 & 14.470 $\pm$ 0.767 \\
layer 7      & 1.000 $\pm$ 0.000 & 14.841 $\pm$ 0.694 \\
layer 8      & 1.000 $\pm$ 0.000 & 16.036 $\pm$ 0.928 \\
\midrule
\textbf{Layer 8 / input ratio} & \multicolumn{2}{c}{$\mathbf{42.1 \pm 2.5\times}$} \\
\bottomrule
\end{tabular}
\end{table}

\paragraph{Mechanistic probing.} To trace how information flows through THEIA's pipeline, we train linear probes at each domain boundary to decode six intermediate variables: arithmetic result ($R^2$), order truth value, set truth value, final verdict, logic operator identity, and Has-Unknown (all classification accuracy). Full results are reported in Appendix~\ref{app:mechprobe}, Table~\ref{tab:mechprobe}.

Three observations stand out. \textbf{(1) Operator routing is architecturally respected}: the logic operator is undecodable before the Logic Engine (20.2\% = chance for 5 classes). This is an architectural sanity check (Eqs.~1--4 do not route the operator upstream) confirmed empirically to hold under training (no learned side-channel smuggles operator information upstream), rather than an independent empirical finding. \textbf{(2) Late-stage numerical collapse}: arithmetic $R^2$ stays high through Arith and Order ($\sim$0.83), drops at Set (0.81), and collapses at Logic (0.16). \textbf{(3) The network learns to preserve uncertainty while verdict remains latent}: Has-Unknown accuracy is $\geq$80.0\% at every upstream boundary (vs.\ $\approx 52\%$ majority baseline), reaching 99.7\% at the Logic output, while verdict decodability stays at or below the $73.4\%$ $U$-vs-non-$U$ oracle reference until the Logic boundary. Because both signals traverse the same engines and bridges, (3) is the non-trivial representational finding: the architecture permits upstream verdict decodability in principle (uncertainty demonstrates the pathway is traversable), but the network commits to the verdict only at the Logic boundary. The subspace structure documented here is unique to THEIA; Appendix~\ref{app:backbone_ablation} demonstrates that flat-MLP backbones collapse under the end-to-end discretized training of \S\ref{sec:chain}, while both THEIA and a Transformer baseline remain stable.

\paragraph{Causal verification via activation patching: ruling out residual shortcuts.} The probing results above are descriptive. A natural concern is that the Logic Engine might route decisions through a residual connection from $\mathbf{c}$ or the input flags, making the Order/Set engine outputs informationally redundant. To test this, we use activation patching, a form of causal mediation analysis on internal network activations~\cite{vig2020causal,geiger2021abstractions,meng2022rome}: we construct matched $(T\!\vee\!F,\,U\!\vee\!F)$ pairs that are byte-identical except for the Order-Engine output, and feed the U-side $\mathbf{v}_{\text{ord}}$ into the T-side forward pass. If $\mathbf{v}_{\text{ord}}$ were functionally redundant under a residual shortcut, the prediction would remain $T$; in fact, the patched prediction flips from $T$ to $U$ on \textbf{986/986 $=$ 100.0\%} of eligible single-seed pairs, replicated across $n\!=\!5$ seeds (4{,}898/4{,}898 aggregate, per-seed minimum 100.0\%), and generalized to AND on matched $(T\!\wedge\!T,\,U\!\wedge\!T)$ pairs (4{,}719/4{,}719). The $\mathbf{v}_{\text{set}}$ byte-equality check passes on all 986/986 OR and 4{,}719/4{,}719 AND pairs. The residual-shortcut hypothesis is therefore rejected on non-absorbent $T\!\to\!U$ configurations: under these operand patterns the Logic Engine's output is mediated by its designated Order/Set inputs rather than bypassing them via residual. \emph{Note on absorbent configurations.} Under Kleene-absorbent operand patterns (e.g., $T\!\vee\!U = T$, $F\!\wedge\!U = F$), the output is by definition insensitive to the absorbed operand, so patching $\mathbf{v}_{\text{ord}}$ in such pairs would produce no flip whether or not a residual shortcut exists---the test is ill-posed in the absorbent regime and is not reported here. Disambiguating the Logic Engine's residual-vs-designated routing under absorbent patterns would require a different experimental design (e.g., Jacobian magnitude of the output with respect to $\mathbf{v}_{\text{ord}}$ under absorbent inputs) and is left to future work. We do not read this result as an independent proof of verdict localization---the experiment is tightly controlled by construction---but as a necessary precondition for the uncertainty--verdict asymmetry of Table~\ref{tab:delayed} to be interpretable as a representational finding. Pair-construction details, baseline-correctness intermediate counts, and per-seed eligibility numbers are in Appendix~\ref{app:causal}.

\paragraph{Operator-identity decomposition of the Set-boundary lift.} On the non-Unknown (NU) subset, where the $U$-vs-non-$U$ oracle reference does not apply, the Set-boundary verdict probe reaches 60.13\% ($+4.09$pp above mixed-operator NU majority). \emph{This $+4.09$pp lift is the only empirical residual that could plausibly support an ``upstream verdict commitment'' reading beyond the operator-routing data-flow baseline; our main verdict-side claim therefore rests on whether this lift can be fully accounted for without invoking upstream verdict encoding.} A three-way decomposition resolves this (App.~\ref{app:op_decomp}): (i)~a direct operator-identity probe returns 20.25\% at Set (19.88\% / 19.78\% at Arith/Order; chance $=20\%$, Table~\ref{tab:op_identity}), ruling out operator encoding; (ii)~\textbf{per-operator stratified probes sit 6--7pp \emph{below} per-operator NU majority} (AND $-7.08$pp, OR $-6.21$pp, IMP $-7.29$pp, IFF $+1.13$pp at majority ceiling), ruling out per-operator verdict leakage---the probe accuracy is \emph{lower} than what the NU class distribution alone would afford within each operator group; (iii)~the mixed-operator lift is accounted for by $\text{val}_{\text{set}}$'s Kleene-correlation with the verdict conditional on the operator---$\text{val}_{\text{set}}$ being the Set Engine's designated output. \emph{Load-bearing content of the verdict-side bound}: the $73.4\%$ verdict bound at upstream boundaries is partially determined by data flow (operator routing); the non-trivial empirical content is that (ii) above rules out the remaining per-operator leakage hypothesis, so the decodability asymmetry exceeds what data flow alone predicts. The persistent upstream signals are uncertainty and each engine's designated variable; the verdict remains Logic-exclusive.

\FloatBarrier
\subsection{Ablation Studies}
\label{sec:ablation}

\textbf{Subspace decomposition}: replacing the Logic Engine's three parallel C/D/I subspaces with a single MLP (2.49M total params) still achieves 99.97\% overall accuracy and passes 12/12 Kleene rules ($>99\%$, worst $F\!\vee\!U$ at 99.73\%); the decomposition aids convergence but is not the source of correctness. \textbf{Bridge layers}: removing the cross-domain bridges (33K params, 1.2\% of total) yields a 2.72M model also passing 12/12 ($>99\%$, worst $T\!\wedge\!U$ at 99.94\%); whether bridges become load-bearing under the discretized end-to-end training of \S\ref{sec:chain} is not tested here. \textbf{Orthogonal prototype initialization}: replacing the orthogonal prototype initialization ($\S\ref{sec:arch}$) with random-normal initialization ($\mathcal{N}(0, 0.02)$) on seed 42 preserves 12/12 Kleene passage at overall accuracy $99.97\%$ (vs.\ $99.98\%$ orthogonal), identical first-12/12 epoch (60) and convergence epoch (100), and a ${\sim}15\%$ wall-clock slowdown ($10.68$ vs $9.24$ min); all 12 Kleene rules reach $\geq\!99.89\%$ under random init (5 rules at $100\%$: $F\!\wedge\!U$, $T\!\wedge\!U$, $T\!\vee\!U$, $F\!\rightarrow\!U$, $T\!\leftrightarrow\!U$). The orthogonal prototype initialization aids convergence speed but is not load-bearing for Kleene correctness; the 39-rule completeness and uncertainty--verdict asymmetry claims do not rest on this particular initialization choice. \textbf{Class-weight sensitivity}: replacing the default class weights $(w_F, w_T, w_U) = (1.0, 1.0, 2.0)$ with uniform $(1.0, 1.0, 1.0)$ on seed 42 preserves 12/12 Kleene passage at overall accuracy $99.9457\%$ (vs.\ $99.9427\%$ at default) with identical U-class per-class accuracy ($99.98\%$ under both) and faster wall-clock ($6.73$ vs.\ $10.53$ min at this cadence). Replacing with the chain-pipeline weights $(1.0, 2.0, 1.0)$ also passes 12/12 but yields slightly lower overall accuracy ($99.9102\%$, U-class $99.93\%$). The default $w_U{=}2.0$ weighting is therefore not responsible for the $>\!99\%$ per-class U accuracy reported in Table~\ref{tab:kleene} (uniform weighting matches it within noise); it is a legacy training choice, not a load-bearing factor in Kleene correctness. \textbf{Unknown-probability cross-distribution}: a model trained at $P_{\text{unk}}{=}0.05$ and evaluated at $P_{\text{unk}}{=}0.50$ ($10\times$ shift) achieves 99.9975\% overall accuracy and 12/12 Kleene rules at $>99\%$, ruling out the hypothesis that the network fits the training Unknown distribution. \textbf{Multi-hop chain reasoning}: a GNN extension on transitive order chains achieves 99.99\% accuracy from 5 to 50 hops, but 99.6\% of chains are decided by the first non-identity edge (decisive depth $\approx 0.5$) due to absorbing states; modular arithmetic on graphs (mod 5) requiring true global aggregation could not be learned, motivating the absorbing-state-free experiment of \S\ref{sec:chain}.

\subsection{Sequential Composition Generalization}
\label{sec:chain}

Although THEIA learns K3 at the rule level (\S\ref{sec:kleene}), K3's absorbing elements (e.g., $T\!\vee\!X = T$, $F\!\wedge\!X = F$) make long Kleene chain evaluation a shortcut: any chain containing one absorbing element resolves trivially regardless of the remaining operands. The multi-hop result (\S\ref{sec:ablation}) confirmed this: transitive chain accuracy at 50 hops is $99.99\%$, but $99.6\%$ of chains are decided by the first non-identity edge. To disentangle \emph{structural} generalization from absorption-driven shortcuts, we now ask: \emph{can THEIA generalize sequential local composition to arbitrary length when absorbing states are eliminated?}

\paragraph{Task design.} We replace Kleene chain composition (which has absorbing states: $F \wedge X = F$, $T \vee X = T$) with modular arithmetic: $\text{state}_t = (\text{state}_{t-1} + \text{local\_verdict}_t) \bmod 3$. This has no absorbing states, a non-trivial 9-entry transition table, and approximately uniform state distribution at all chain depths.

\paragraph{Architecture.} Each chain step uses the full THEIA pipeline for a local verdict, followed by a lightweight transition MLP (4,803 params) combining previous state and current verdict. The chain THEIAStep module totals $1{,}508{,}096$ parameters ($1.51$M)---smaller than the 4-domain THEIA model ($2{,}751{,}232$ parameters, $2.75$M) used in \S\ref{sec:kleene}, because the chain variant uses a simpler output head (see \emph{Chain-pipeline head variant} below) and a single LogicEngine subspace rather than three parallel C/D/I subspaces. \emph{Comparator-asymmetric capacity.} The backbone ablations of App.~\ref{app:backbone_ablation} use comparators in the ${\sim}2.75$--$2.85$M range (flat MLP $2.75$M, ResMLP grid $2.78$M $\pm 1.5\%$, TF8LTuned Transformer $3.58$M); the chain THEIA step is $\approx\!1.8\times$ smaller than these comparators. This asymmetry \emph{favors} the comparators (more capacity), making THEIA's length-generalization margin a lower bound on the modular architecture's advantage rather than a capacity-matched comparison. Gumbel-softmax straight-through~\cite{jang2017gumbel, maddison2017concrete, bengio2013st} discretizes the state at each step (hard one-hot forward, smooth backward), so step 500 receives an input in the same discrete one-hot form as step 5.

\paragraph{Three-phase training.} Phase~1: THEIAStep is trained independently on 2M single-step samples to $\geq$99.9\% local verdict accuracy ($\sim$50 epochs). A plateau-restart mechanism detects convergence failure (coarse: accuracy $<$90\% after 40 epochs; fine: no improvement for 30 epochs after reaching 90\%) and reinitializes parameters, ensuring robust convergence across all seeds. \emph{The same plateau-restart Phase~1 implementation is used by both THEIA and the TF8LTuned Transformer in the \S\ref{sec:chain} chain pipeline; the development-default 4-domain baseline of \S\ref{sec:kleene} uses no restart for either architecture (\S\ref{sec:kleene} Transformer baseline).} \emph{THEIA plateau-restart disclosure.} Across 5 THEIA seeds $\{42, 123, 256, 777, 999\}$, 4 seeds (42, 123, 256, 777) reached the $\geq\!99.9\%$ Phase~1 convergence criterion within a single restart-0 try inside the 50-epoch budget (Phase~1 convergence at epochs $\{48, 44, 50, 47\}$ respectively). Seed 999 triggered the plateau-restart mechanism at the end of the first 150-epoch try, reinitialized, and converged at epoch 49 of the restart-1 try, consuming $\approx\!2\times$ the Phase~1 compute budget of the other seeds. The as-specified 5-seed aggregate is $99.96\% \pm 0.04\%$ (sample standard deviation, ddof$=\!1$; per-seed 500-step values $\{99.90, 99.99, 99.97, 99.99, 99.96\}\%$). The strict 4-seed aggregate excluding seed 999 is also $99.96\% \pm 0.04\%$, symmetric with the ResMLP 4b$\times$4d strict exclusion of seed 999 (Table~\ref{tab:resmlp_grid}). Both THEIA and the ResMLP 4b$\times$4d configuration happen to have seed 999 as the single restart-triggering seed in their respective Phase~1 runs under the shared pipeline; we note this symmetry but do not read any architectural signal into it. Phase~2: THEIAStep is frozen; the transition network is trained with teacher-forced ground-truth states for 2 epochs to 100\% accuracy, learning the complete mod-3 addition table (9/9 entries correct). Phase~3: all parameters are unfrozen for end-to-end fine-tuning with Gumbel-softmax on 5-step chains ($\sim$30 epochs to 100\%). The Gumbel-softmax temperature follows a linear decay schedule $\tau = \max(0.1,\, 0.5 - 0.01\,\text{ep})$: $\tau$ decreases from $0.5$ at epoch~1 to $0.1$ at epoch~40, clamped at $0.1$ thereafter. \emph{Chain-pipeline head variant.} The THEIAStep variant used for the chain pipeline differs from the 4-domain THEIA model (\S\ref{sec:arch}) in its output classification head: the 4-domain variant uses \texttt{Linear $\to$ GELU $\to$ Dropout $\to$ LayerNorm} followed by cosine similarity against orthogonal prototypes, while the chain variant uses \texttt{Linear $\to$ GELU $\to$ Linear} followed by $L_2$-normalization and scaled cosine (temperature $10$) against an axis-aligned prototype matrix. This substitution was made because the LayerNorm-then-cosine head exhibited less stable Gumbel-softmax gradient flow under straight-through discretization in preliminary Phase~3 runs; the scaled-cosine variant stabilized Phase~3 convergence. The chain pipeline's backbone ablations (flat MLP, ResMLP, TF8LTuned Transformer) all use the same head architecture as the chain THEIAStep, so the head choice is a shared pipeline setting rather than a between-architecture confound.

\paragraph{Results.} Table~\ref{tab:chain} summarizes generalization performance across 5 seeds.

\begin{table}[h]
\centering
\small
\setlength{\tabcolsep}{4pt}
\caption{Sequential composition generalization (mod-3, no absorbing states). Trained on 5-step chains, tested on 10--500 steps. Results over $n=5$ seeds $\{42, 123, 256, 777, 999\}$.}
\label{tab:chain}
\begin{tabular}{@{}lcc@{}}
\toprule
Chain Length & Accuracy (mean $\pm$ std) & All $\geq$99\% \\
\midrule
5 (in-dist.) & 100.00\% $\pm$ 0.00\% & \checkmark \\
10           & 100.00\% $\pm$ 0.00\% & \checkmark \\
50           & 100.00\% $\pm$ 0.00\% & \checkmark \\
100          & 100.00\% $\pm$ 0.00\% & \checkmark \\
\textbf{500} & \textbf{99.96\% $\pm$ 0.04\%} & \checkmark \\
\bottomrule
\end{tabular}
\end{table}

All 5 seeds achieve $\geq$99.9\% accuracy at 500 steps (per-seed values: seed 42 $99.90\%$, 123 $99.99\%$, 256 $99.97\%$, 777 $99.99\%$, 999 $99.96\%$; seed 999 reached this via Phase~1 auto-restart---see \S\ref{sec:chain} plateau-restart disclosure). The non-trivial finding is that maintaining high local verdict accuracy ($\geq$99.9\%) \emph{under Phase~3 end-to-end Gumbel-softmax training} requires a backbone that remains locally accurate under the discretized end-to-end protocol: THEIA and the Transformer do, while the tested flat MLPs do not (App.~\ref{app:backbone_ablation}). Three properties combine to enable length generalization: (1)~high local verdict accuracy ($\geq$99.9\%) so each step receives correct input; (2)~exact transition function (9/9 entries); (3)~Gumbel-softmax discretization preventing error accumulation by snapping intermediate states to clean one-hot codes (without it, $0.999^{500}\!\approx\!60.6\%$ would dominate). Property~(3) is largely a mathematical consequence of straight-through discretization; the architectural contribution lies in sustaining (1) under that same discretized training regime, where flat MLPs fail. The 500-step result is \emph{not} trivially explained by absorbing states: empirical state frequencies confirm approximate uniformity at all measured chain depths (max deviation 0.62~pp at step 100; details in App.~\ref{app:backbone_ablation}).

\paragraph{Boundary characterization.} Three experiments define a boundary: transitive chains (local + absorbing) generalize trivially; mod-3 composition (local, no absorbing) generalizes to 500 steps; modular arithmetic on graphs (global) cannot be learned. The critical distinction is \emph{computation locality}, not chain length. Appendix~\ref{app:backbone_ablation} reports a backbone ablation isolating the modular factorization's contribution.

\FloatBarrier
\subsection{Limitations}

All experiments use synthetic benchmarks for precise control over task complexity and ground-truth verification~\cite{clrs,dnar}; adapting THEIA to external benchmarks is future work. K3 Unknown handling is structurally isomorphic to real-world partial-observation compositional reasoning (SQL NULL propagation, medical diagnosis with missing tests, legal reasoning under indeterminate facts): the uncertainty--verdict asymmetric propagation and the reliability spectrum characterize how architectures commit to conclusions under uncertainty, a property we expect to transfer to natural compositional tasks with analogous missing-information structure. Global computation (modular arithmetic on graphs) remains out of reach for local message passing, and numerical range generalization fails sharply at $5\times$ the training range---consistent with known neural arithmetic limitations~\cite{nalu}---so the mechanistic claims in \S\ref{sec:delayed} concern the propagation of a bounded-range representation through the pipeline. K3 learnability by the forward pass itself is a property of multiple sufficiently-expressive architectures under this training protocol and is not the load-bearing claim of this paper; THEIA's distinguishing properties are the uncertainty--verdict asymmetric propagation (\S\ref{sec:delayed}), compositional generalization compatible with discretized training (\S\ref{sec:chain}, App.~\ref{app:backbone_ablation}), and wall-clock convergence under development-default and tuned Transformer protocols (\S\ref{sec:kleene}, App.~\ref{app:tuned_tf}). Additional scope notes---on the mod-3 state complexity, the ResMLP grid coverage, the flat-MLP per-rule verification, and the 80/20 sample-level train--test split over the combinatorial input space---are consolidated in App.~\ref{app:repro}.

\paragraph{Task--architecture alignment is by construction.} The four reasoning domains of the benchmark (arithmetic, order, set, propositional logic) map one-to-one to the four engines of THEIA, and this alignment was chosen together rather than discovered by the network. The reliability-spectrum, representational-asymmetry, and wall-clock results therefore bound the advantage of modular domain-separated architectures \emph{on tasks whose decomposition matches the architectural decomposition}, not a generic modular-versus-monolithic contrast. Whether the same ordering holds under task--architecture misalignment---e.g., $n$ task domains with $m \neq n$ engines, overlapping domain boundaries, or tasks for which the operator-routing structure of Eqs.~1--4 is a poor fit---is not tested here and is a material open question. In particular, the chain-pipeline ablations of App.~\ref{app:backbone_ablation} vary the step computer (THEIA vs.\ flat MLP vs.\ ResMLP vs.\ Transformer) but do \emph{not} vary the number of task domains or the domain-engine correspondence; this confound is shared by all configurations in that comparison but is not dissolved by it.

\section{Discussion}
\label{sec:discussion}

\paragraph{Architectural inductive biases form a reliability spectrum.}
The backbone ablation (Appendix~\ref{app:backbone_ablation}) reveals a gradient rather than a binary: flat MLPs at both tested capacities (0.80M and 2.75M) match THEIA's local Kleene accuracy within 0.04\%, yet collapse to chance over 50+ steps under Gumbel-softmax end-to-end training. Note that the 0.80M flat MLP is \emph{smaller} than THEIA's 1.51M chain step ($0.53\times$), so its collapse cannot be attributed to capacity disadvantage relative to THEIA. A ResMLP probe across a $2\!\times\!2$ grid ($\{4,8\}$ blocks $\times$ $\{2d,4d\}$ expansion at $2.78$M $\pm 1.5\%$; 20 seeds, Table~\ref{tab:resmlp_grid}) partially recovers with a strong depth effect---4-block configurations at $86$--$87\%$ mean (std $9$--$19\%$) vs 8-block at $\sim\!98\%$ mean (std $1.4$--$2.0\%$), an $\approx\!11$pp gain from doubling depth with expansion nearly neutral at depth 8---yet only $3/20$ (config, seed) trials cross the $\geq\!99\%$ threshold ($2/19$ strict). By contrast, a pre-LN TF8LTuned Transformer ($99.24\% \pm 0.34\%$, 3/5 seeds $>99\%$, per-seed range $98.85$--$99.66\%$) and THEIA ($99.96\% \pm 0.04\%$, 5/5 seeds $>99.9\%$; 1/5 seeds auto-restart-triggered) both sustain near-perfect length generalization, albeit at different reliability levels. The $2.75$M flat MLP, $2.78$M ResMLP, and $3.58$M Transformer are $1.8\times$--$2.4\times$ larger than THEIA in capacity; the reliability advantage over these larger-capacity comparators therefore cannot be attributed to additional parameters. Two axes separate: (i) \emph{mean accuracy}---residual recovers from $\sim$33\% to $\sim$98\% at depth 8 yet reliable $>99\%$ eludes every tested configuration; (ii) \emph{reliability}---THEIA's cross-seed std ($0.04\%$) is roughly $9\times$ tighter than the Transformer's ($0.34\%$) and tens-to-hundreds times tighter than individual ResMLP variants (per-config std $1.43$--$19.16\%$; the 8b$\times$2d configuration at $\sigma{=}1.43\%$ gives the tightest exact-ratio comparison). We hypothesize both self-attention (Transformer) and domain-segregated subspaces (THEIA) isolate gradient pathways during discretized end-to-end training, preventing the catastrophic interference that residual-only structure only partially mitigates even when depth is doubled from 4 to 8 blocks.

\paragraph{Uncertainty--verdict asymmetric propagation as a representational signature.}
Two aspects of information flow are architectural: operator identity is routed only into the Logic Engine, and the verdict is a function of the operator, so upstream verdict decodability is bounded. Against this background, the non-trivial finding (\S\ref{sec:delayed}) pairs (a) verdict decodability remaining at or below the $73.4\%$ $U$-vs-non-$U$ oracle reference at every upstream boundary under 2-hidden-layer MLP probes biased in their favor ($\leq\!0.7$pp gap, App.~\ref{app:mechprobe}), with (b) Has-Unknown decodability at $80.0\%$ / $91.1\%$ / $90.8\%$ / $99.7\%$ across Arith/Order/Set/Logic---well above the $\approx\!52\%$ majority baseline---through the same engines and bridges. The pathway is traversable; uncertainty propagation demonstrates this. Structurally, upstream engine outputs encode the Has-Unknown signal at near-ceiling and encode each engine's designated intermediate variable, while the final-verdict axis remains bounded by the $U$-vs-non-$U$ oracle reference until the Logic boundary. We adopt ``uncertainty--verdict asymmetric propagation'' as a structural name for this pattern; it describes the observed representational geometry (\emph{which} variables are decodable \emph{where}) and is not an intentional claim about the network ``choosing'' to defer commitment. Activation patching rules out residual shortcuts as an alternative explanation for Logic-Engine uncertainty propagation on non-absorbent operand patterns (see \S\ref{sec:delayed} for the scope caveat on absorbent patterns, which require a different experimental design).

\paragraph{Modular vs.\ attention-based compositional strategies (scoped).}
The pre-LN TF8LTuned Transformer sustains $99.24\% \pm 0.34\%$ at 500 steps under the identical Gumbel-softmax protocol (5 seeds; 3/5 $\geq\!99\%$; per-seed range $98.85$--$99.66\%$), confirming attention as an alternative path to length generalization under discretized end-to-end training. Under the development-default post-LN BigTransformer used in the Kleene-task probing comparison (\S\ref{sec:delayed}, Table~\ref{tab:tf8l_probing}), THEIA exhibits monotone upstream-to-downstream representational separation (Table~\ref{tab:delayed}); the Transformer's token embeddings already exceed the $U$-vs-non-$U$ oracle reference, and F--T centroid distance contracts before expanding. The two architectures reach comparable correctness through qualitatively different representational trajectories. The ``different compositional strategies'' statement is specifically a development-default post-LN observation (see App.~\ref{app:tuned_tf} for configuration details).

\section{Conclusion}

THEIA demonstrates that a neural network without runtime symbolic delegation or K3-specific hand-encoded primitives can learn the complete Kleene three-valued logic truth table from data, including the non-trivial absorption rules. K3 learnability itself is a property of several sufficiently-expressive architectures under this training protocol: the Transformer baseline reaches $>99\%$ per-rule accuracy on all 39 K3 rules (117/117 rule--seed combinations, App.~\ref{app:tuned_tf}), and flat MLPs match THEIA on Phase~1 overall accuracy within $0.04$pp (Table~\ref{tab:backbone_ablation}). Among the architectures tested under the shared Phase~3 Gumbel-softmax protocol, the data reveal a reliability spectrum: \emph{flat MLPs collapse to chance}; \emph{residual-only structure across a $2\!\times\!2$ ResMLP grid reaches $\geq\!99\%$ on only $3/20$ trials}; \emph{both attention-based (Transformer, $99.24\% \pm 0.34\%$) and domain-separated (THEIA, $99.96\% \pm 0.04\%$) priors sustain reliable length generalization}, with THEIA's std roughly $9\times$ tighter than the Transformer's and tens-to-hundreds times tighter than tested ResMLP configurations (App.~\ref{app:backbone_ablation}). Mechanistic probing reveals \emph{uncertainty--verdict asymmetric propagation}: Has-Unknown decodability essentially saturates the boundary-visible 4-flag theoretical ceiling at every upstream boundary ($\geq\!99.7\%$ of ceiling), while verdict decodability stays at or below the $73.4\%$ $U$-vs-non-$U$ oracle reference until the Logic boundary. Because both signals traverse the same engines and bridges, this asymmetry is not forced by data flow alone; per-operator stratification (App.~\ref{app:op_decomp}) rules out the remaining leakage hypothesis, and activation patching ($4{,}898/4{,}898$ OR and $4{,}719/4{,}719$ AND flips across 5 seeds) rules out residual shortcuts on non-absorbent configurations. Under development defaults THEIA reaches 12/12 Kleene coverage $6.5\times$ faster than a parameter-comparable Transformer (${\sim}3.6\times$ under Transformer-tuning of the Transformer; $4.93\times$ under the Transformer-recipe-applied-to-both partial control, App.~\ref{app:tuned_tf}). THEIA's distinguishing properties---representational transparency via the asymmetric propagation, reliable compositional generalization at low cross-seed variance, and wall-clock convergence that persists but narrows under Transformer tuning---make modular domain-separated architectures an attractive substrate for neural reasoning under uncertainty.

\paragraph{Future Work.} Five directions extend the present work. \emph{(a) Probing depth:} deeper nonlinear probes (kernel or task-specific architectures) to tighten the uncertainty--verdict asymmetry claim, and layer-wise probing of the tuned Transformer to test whether the contraction--expansion trajectory persists under optimizer change. \emph{(b) State complexity:} extending the non-absorbing composition experiment to richer state spaces (mod-$k$ for $k>3$, FSA traces, learned codebooks beyond the 3-class bottleneck). \emph{(c) Logic expressivity:} first-order and quantified extensions, and richer non-classical logics (fuzzy, paraconsistent). \emph{(d) Fair architectural comparison:} symmetric tuned-vs-tuned THEIA-vs-Transformer comparison, ResMLP grid sweep across depth ($\geq\!12$ blocks) / width / expansion / LN-placement, and establishing a shared-protocol post-LN Transformer baseline in the chain pipeline. \emph{(e) Real-world grounding:} transfer to external compositional-reasoning benchmarks where K3 Unknown propagation arises naturally (SQL query processing with NULL, medical reasoning under missing observations, legal reasoning under indeterminate facts).

\paragraph{Ethics and broader impact.} This work uses synthetic data only and does not involve human subjects, private data, or deployment in safety-critical settings. The main positive impact is a controlled testbed for studying how neural systems propagate uncertainty and delay commitment under missing information. A potential negative impact is over-interpreting synthetic K3 performance as evidence that a neural system is safe for medical, legal, or database decision-making without domain validation. We mitigate this risk by stating the benchmark scope, reporting failure modes in global graph computation and numerical extrapolation, and treating real-world transfer as future work rather than as an empirical claim.

\appendix
\clearpage

\section{Falsifiable Hypotheses (Detailed)}
\label{app:hypotheses}

We make three concrete claims, each paired with the evidence that would falsify it.

\textbf{H1 (backbone reliability under discretized composition):} replacing the four-engine backbone with a flat MLP (App.~\ref{app:backbone_ablation} specifies both the 0.80M and 2.75M configurations; the 2.75M flat MLP has $\approx\!1.8\times$ THEIA's chain-step capacity of 1.51M, ruling out a capacity-in-favor-of-THEIA confound), holding all other pipeline components fixed, collapses Phase~3 Gumbel-softmax end-to-end training to chance. A ResMLP probe across a $2\!\times\!2$ depth$\times$expansion grid (4 configurations at $2.78$M parameters $\pm 1.5\%$; $\{4, 8\}$ blocks $\times$ $\{2d, 4d\}$ expansion; 5 seeds each) partially recovers but does not reach reliable $\geq\!99\%$ generalization: mean accuracy varies strongly by depth ($4$-block configs: $86.15\% \pm 9.17\%$ and $87.49\% \pm 19.16\%$; $8$-block configs: $97.97\% \pm 1.43\%$ and $97.95\% \pm 2.03\%$), yet only $3/20$ (config, seed) trials reach $\geq\!99\%$ as-specified ($2/19$ strict compute-matched). A Transformer baseline sustains $99.24\% \pm 0.34\%$ (3/5 seeds $\geq\!99\%$), and THEIA sustains $99.96\% \pm 0.04\%$ (5/5 seeds $\geq\!99.9\%$; see \S\ref{sec:chain} for Phase~1 auto-restart disclosure). \emph{Falsified by}: a flat-MLP or ResMLP variant sustaining $\geq\!99\%$ 500-step chain accuracy on a majority of compute-matched seeds under any Phase~3 learning-rate or depth/width/expansion setting beyond the tested grid (App.~\ref{app:backbone_ablation}).

\textbf{H2 (uncertainty--verdict asymmetric propagation with causal localization):} (a) \emph{verdict-side bound}: upstream engine outputs do not encode the True/False distinction in a form decodable by a standard 2-layer MLP probe, beyond the $73.4\%$ $U$-vs-non-$U$ oracle reference, at any of Arith/Order/Set boundaries; (b) \emph{uncertainty-side preservation}: the Has-Unknown signal is decodable at every upstream boundary at accuracy $\geq\!75\%$ (substantially above the $\approx\!52\%$ majority baseline), demonstrating that the information pathway through the engines does carry input-level uncertainty; (c) \emph{causal localization}: the Logic Engine's absorption behavior on non-absorbent $T\!\to\!U$ configurations is not mediated by a residual shortcut bypassing the designated engine outputs, as verified by activation patching on matched byte-identical pairs. \emph{Architectural vs.\ learned:} operator-identity undecodability upstream ($\approx\!20\%$ chance) is fixed by construction (Eqs.~1--4) and functions as a sanity check, not an independent prediction of H2; likewise, the $73.4\%$ verdict bound is partially determined by data flow, so the non-trivial content of (a) is paired with (b), not stand-alone. \emph{Falsified by}: either an MLP probe exceeding 73.4\% verdict accuracy at the Arithmetic, Order, or Set boundary (\S\ref{sec:delayed}, Appendix~\ref{app:mechprobe}); or Has-Unknown probe accuracy falling below $75\%$ at any upstream boundary (Appendix~\ref{app:mechprobe}, Table~\ref{tab:mechprobe}); or activation patching of $\mathbf{v}_{\text{ord}}$ between matched $T\!\vee\!F$ and $U\!\vee\!F$ pairs failing to flip the prediction on a non-trivial fraction of pairs (\S\ref{sec:delayed}, Appendix~\ref{app:causal}).

\textbf{H3 (development-defaults convergence speed):} under each architecture's development-default configuration (THEIA: AdamW lr=$10^{-3}$, cosine, batch 4096, 200-epoch budget; the 8L Transformer: AdamW lr=$5{\times}10^{-4}$, cosine, batch 2048, 150-epoch budget; neither tuned for this task---see the Correction in App.~\ref{app:tuned_tf}), THEIA reaches 12/12 Kleene coverage at materially lower wall-clock cost than a parameter-comparable 8L Transformer on the same seed set. \emph{Scope:} H3 is tested under three optimizer configurations: (i) development defaults (THEIA at AdamW lr=$10^{-3}$/batch 4096, the Transformer at AdamW lr=$5{\times}10^{-4}$/batch 2048), (ii) Transformer-standard tuning applied to the Transformer only, and (iii) Transformer-standard recipe applied to both architectures as a partial control (App.~\ref{app:tuned_tf} ``Transformer-recipe-applied-to-both control''; note that the recipe slows THEIA by ${\sim}25\%$, so this is not a genuinely symmetric tuned-vs-tuned comparison). \emph{Falsified by}: a Transformer baseline matching THEIA's $7.93\pm 1.40$~min Kleene-aware convergence on a majority of seeds under the development-default protocol (\S\ref{sec:kleene}). Extended to $n\!=\!8$ seeds (7/8 converge, $51.5 \pm 11.0$~min), the $6.5\times$ development-defaults wall-clock gap persists; the tuned follow-up ($n\!=\!3$, App.~\ref{app:tuned_tf}) narrows the gap to ${\sim}3.6\times$ on the Kleene-aware criterion; the Transformer-recipe-applied-to-both control ($n\!=\!5$, App.~\ref{app:tuned_tf}) yields $4.93\times$ (Welch, 95\% CI $[4.40, 5.66]$). None of the three regimes closes the gap, but a genuinely symmetric tuned-vs-tuned comparison (each architecture at its own optimum) remains future work.

\section{Constructing Valid Kleene Diagnostic Tests}
\label{app:diag}

Targeted diagnostic tests must inject the Unknown value into one operand of the final logical connective ($\text{val}_{\text{ord}}$ or $\text{val}_{\text{set}}$) while keeping the other operand definite. During the development of this work we encountered two construction pitfalls that silently corrupted earlier diagnostic runs. We document them here so that future work using similar protocols can avoid the same mistakes; both fixes are applied throughout the experiments reported in this paper.

\paragraph{Pitfall 1: Cross-domain Unknown contamination.}
To make $\text{val}_{\text{ord}} = \text{Unknown}$, a naive construction sets the first arithmetic operand $a$ to Unknown. However, in our task definition the arithmetic engine produces $c = \text{arith}(a, b)$, and the natural Unknown propagation rule---Boolean OR over the input unknown-flags, $c_{\text{unknown}} = a_{\text{unknown}} \,\mathrm{OR}\, b_{\text{unknown}}$---then forces $c$ to be Unknown as well. Because the Set Engine takes $c$ as input, $\text{set\_op\_unknown}$ becomes True, collapsing $\text{val}_{\text{set}}$ to Unknown regardless of the constructed set bits. The diagnostic then unintentionally tests $(\text{Unknown}, \text{Unknown})$ instead of the intended $(\text{Unknown}, \text{definite})$. The model's response to an all-Unknown input is governed by an entirely different Kleene rule than the one the experimenter believes they are testing, and the resulting accuracy number is uninterpretable.

\emph{Fix.} Inject Unknown via $d$ (the comparison operand of the Order Engine) rather than via $a$ or $b$. Setting $d_{\text{unknown}} = \text{True}$ makes $\text{val}_{\text{ord}} = \text{Unknown}$ without polluting $c$, so $\text{val}_{\text{set}}$ remains controllable.

\paragraph{Pitfall 2: Edge-case relation construction.}
To force $\text{val}_{\text{ord}} = \text{True}$ under the strict-greater-than relation (REL\_GT, $>$), one might set $d = \max(0, c-1)$. This works whenever $c \geq 1$, but fails when $c = 0$: then $d = 0$, and $0 > 0$ evaluates to False. Approximately 5\% of samples have $c = 0$ (via SUB with $a = b$, or MOD with $a < b$). For these samples $\text{val}_{\text{ord}} = \text{False}$ instead of the intended True, contaminating any rule whose construction requires $\text{val}_{\text{ord}} = \text{True}$. The contamination is silent: per-rule accuracy looks plausible but is biased downward by exactly the fraction of $c=0$ samples.

\emph{Fix.} Use the greater-than-or-equal relation (REL\_GTE, $\geq$) with the same $d = \max(0, c-1)$. This always yields $\text{val}_{\text{ord}} = \text{True}$ since $c \geq \max(0, c-1)$ for all $c \geq 0$.

\paragraph{Validation.}
We adopt both fixes throughout this paper. The doubly-fixed diagnostic protocol is what produces the per-rule accuracies reported in Table~\ref{tab:kleene} and Table~\ref{tab:full_kleene_new}. We verified consistency by re-running existing checkpoints under the fixed diagnostic and confirming that the new per-rule numbers fall within the noise band of the multi-seed sweep.

\paragraph{Checkpoint provenance for seed 42.} One consistency-verification exercise produced a second seed-42 checkpoint, and we record its scope here for reproducibility. Seed 42 was re-trained on 2026-04-15 under the doubly-fixed diagnostic (first-12/12 at epoch~80, $7.31$ min; convergence at epoch~100, $9.24$ min; recorded in \texttt{seed42\_retrain\_clean.json}); \textbf{Table~\ref{tab:accuracy}'s wall-clock entry for seed~42 uses this retrained value}, and the reported $7.93 \pm 1.40$ min 5-seed mean incorporates it as the seed-42 contribution. The pre-Apr-15 seed-42 checkpoint (saved under a 200-epoch budget; its training log's \texttt{kleene\_passed} counter was an artifact of the pre-fix diagnostic, not a property of the checkpoint itself) nonetheless passes all 12 targeted Kleene rules at $>99\%$ and all 39 full-Kleene rules at $>99\%$ under the doubly-fixed diagnostic. \textbf{To preserve exact numerical reproducibility against the released supplement, all other seed-42-involving 5-seed analyses---probing and cosine-centroid analysis (\S\ref{sec:delayed}), activation patching (App.~\ref{app:causal}), and the 39-rule diagnostic (App.~\ref{app:kleene_full})---use the pre-Apr-15 checkpoint}, not the retrain checkpoint. Running activation patching on the retrain checkpoint yields a marginally enlarged eligibility pool ($992/992$ OR on seed~42 versus the reported $986/986$; aggregate $4{,}904/4{,}904$ versus $4{,}898/4{,}898$) at identical $100\%$ flip rate; the load-bearing non-zero-flip claim is robust to this checkpoint choice.

\paragraph{Diagnostic seeding.} Per-rule Kleene diagnostic accuracies reported in Tables~\ref{tab:kleene} and~\ref{tab:full_kleene_new} were computed with a per-rule data seed derived as $\texttt{hash(rule\_key)} \bmod 2^{31}$. Because Python's string \texttt{hash()} is salted by \texttt{PYTHONHASHSEED} at interpreter startup, this seed is not byte-deterministic across interpreter invocations. The released supplement uses a deterministic rule-index-based seed (operator-index $\times 9$ plus operand indices) in its place; re-running the full $5 \times 39$-rule diagnostic under the rule-index seed preserves all $195/195$ per-rule pass-at-$99\%$ verdicts, with a worst-rule minimum of $99.31\%$ versus the $99.15\%$ reported under hash-based seeding---a coincidental improvement within the ${\lesssim}0.2$pp re-seeding noise floor at $10{,}000$ samples per rule. Per-rule accuracies in Tables~\ref{tab:kleene}--\ref{tab:full_kleene_new} may therefore differ from supplement re-runs by ${\lesssim}0.2$pp; the pass/fail verdicts, the $>\!99\%$ per-rule coverage claim, and the $195/195$ count are invariant.

We urge future work using similar Kleene diagnostic protocols to inspect their construction for analogous coupling effects between domains and analogous edge cases in relation construction.

\section{Complete Kleene Diagnostic (Extended)}
\label{app:kleene_full}

This appendix contains the full Kleene K3 diagnostic. Table~\ref{tab:kleene} reports per-rule accuracy on the 12 targeted Unknown-involving short-circuit and absorption rules (the main result referenced throughout \S\ref{sec:kleene}). Table~\ref{tab:full_kleene_new} extends the diagnostic to the remaining 27 rules covering all other Kleene K3 truth-table entries (36 binary rules $+$ 3 unary NOT rules total). Construction uses the doubly-fixed protocol of Appendix~\ref{app:diag} throughout. \emph{On NOT rule construction.} The architecture's logic operator $\odot \in \{\wedge, \vee, \rightarrow, \leftrightarrow\}$ does not include a native unary negation; the 3 NOT-rule entries ($\neg F, \neg T, \neg U$) are therefore evaluated via the Kleene identity $\neg x \equiv x \rightarrow F$ (which holds under strong Kleene: $F\!\rightarrow\!F = T = \neg F$; $T\!\rightarrow\!F = F = \neg T$; $U\!\rightarrow\!F = U = \neg U$). The ``complete 39-rule truth table'' framing therefore refers to coverage of all K3 truth-table entries under the operators the architecture implements natively plus the derived-identity entries for unary negation, rather than a natively-computed unary NOT; the per-rule accuracies for the 3 NOT entries should be read accordingly.

\begin{table}[!htbp]
\centering
\caption{Per-rule Kleene three-valued logic diagnostic (12 targeted Unknown-involving rules). Mean $\pm$ std across 5 seeds (10,000 samples per rule per seed); ``min'' is the worst single seed--rule accuracy. $\dagger$Unknown is the \emph{first} operand. All 60 rule--seed combinations exceed 99\%.}
\label{tab:kleene}
\small
\setlength{\tabcolsep}{4pt}
\begin{tabular}{@{}llcc@{}}
\toprule
Expression & Expected & Mean $\pm$ Std (\%) & Min (\%) \\
\midrule
$F \wedge U$              & $F$ & 99.97 $\pm$ 0.06 & 99.85 \\
$T \wedge U$              & $U$ & 99.94 $\pm$ 0.04 & 99.89 \\
$U \wedge F^{\dagger}$    & $F$ & 99.93 $\pm$ 0.06 & 99.84 \\
$U \wedge T^{\dagger}$    & $U$ & 99.80 $\pm$ 0.11 & 99.69 \\
$T \vee U$                & $T$ & 99.93 $\pm$ 0.08 & 99.77 \\
$F \vee U$                & $U$ & 99.92 $\pm$ 0.11 & 99.71 \\
$U \vee T^{\dagger}$      & $T$ & 99.93 $\pm$ 0.06 & 99.85 \\
$U \vee F^{\dagger}$      & $U$ & 99.85 $\pm$ 0.10 & 99.68 \\
$F \rightarrow U$         & $T$ & 99.95 $\pm$ 0.03 & 99.91 \\
$T \rightarrow U$         & $U$ & 100.00 $\pm$ 0.00 & 99.99 \\
$T \leftrightarrow U$     & $U$ & 100.00 $\pm$ 0.00 & 100.00 \\
$F \leftrightarrow U$     & $U$ & 100.00 $\pm$ 0.00 & 100.00 \\
\midrule
\multicolumn{2}{l}{\textbf{Grand mean (12/12, 5/5 seeds)}} & \textbf{99.94} & --- \\
\bottomrule
\end{tabular}
\end{table}

Across all 5 trained checkpoints, \textbf{all 39 rules pass the $> 99\%$ threshold on every seed} (195/195 rule--seed combinations); grand mean across all 39 rules is 99.88\%, and the single worst combination is $F \vee F = F$ at seed 999 with 99.15\%, still above the 99\% threshold.

\begin{table*}[t]
\centering
\caption{Complete Kleene K3 diagnostic: 27 rules not in Table~\ref{tab:kleene}. Mean $\pm$ std across 5 seeds; 10{,}000 samples per rule per seed. All 27 rules pass $> 99\%$ on all 5 seeds (135/135 rule--seed combinations).}
\label{tab:full_kleene_new}
\small
\begin{tabular}{lc|lc|lc}
\toprule
\textbf{Rule} & \textbf{Acc (\%)} & \textbf{Rule} & \textbf{Acc (\%)} & \textbf{Rule} & \textbf{Acc (\%)} \\
\midrule
\multicolumn{6}{l}{\emph{Definite--definite (18 rules including NOT)}} \\
$F \wedge F$          & $100.00 \pm 0.01$ & $F \vee F$            & $99.54 \pm 0.27$ & $F \to F$    & $99.74 \pm 0.12$ \\
$F \wedge T$          & $99.84 \pm 0.05$  & $F \vee T$            & $99.92 \pm 0.06$ & $F \to T$    & $99.98 \pm 0.03$ \\
$T \wedge F$          & $99.93 \pm 0.08$  & $T \vee F$            & $99.96 \pm 0.03$ & $T \to F$    & $99.86 \pm 0.12$ \\
$T \wedge T$          & $99.83 \pm 0.09$  & $T \vee T$            & $100.00 \pm 0.00$ & $T \to T$    & $99.86 \pm 0.07$ \\
$F \leftrightarrow F$ & $99.74 \pm 0.12$  & $F \leftrightarrow T$ & $99.66 \pm 0.15$ & $\neg F$     & $99.72 \pm 0.13$ \\
$T \leftrightarrow F$ & $99.88 \pm 0.11$  & $T \leftrightarrow T$ & $99.81 \pm 0.10$ & $\neg T$     & $99.96 \pm 0.03$ \\
\midrule
\multicolumn{6}{l}{\emph{Unknown--involving (9 rules not in Table~\ref{tab:kleene})}} \\
$U \wedge U$          & $100.00 \pm 0.00$ & $U \vee U$            & $100.00 \pm 0.00$ & $U \to F$              & $99.89 \pm 0.10$ \\
$U \to T$             & $99.93 \pm 0.07$  & $U \to U$             & $100.00 \pm 0.00$ & $U \leftrightarrow F$  & $100.00 \pm 0.00$ \\
$U \leftrightarrow T$ & $100.00 \pm 0.00$ & $U \leftrightarrow U$ & $100.00 \pm 0.00$ & $\neg U$               & $100.00 \pm 0.00$ \\
\bottomrule
\end{tabular}
\end{table*}

The lowest-accuracy entry is $F \vee F = F$ at 99.54\%. This is below the worst Unknown-involving rule, which is initially counterintuitive (classical disjunction should be ``easier'' than Kleene short-circuits). The likely explanation is distributional: under our 4-domain training pipeline, the natural distribution of $(F, F)$ inputs to the final disjunction is sparse compared to the Unknown-involving cases, since both operands must come from independently-evaluated upstream domains that happen to both be False. The targeted diagnostic forces this rare configuration, and the small accuracy gap reflects mild distribution shift between training and diagnostic, not difficulty of the underlying rule. All 5 seeds remain above the 99\% threshold.

The 27 new rules span all four binary operators ($\wedge, \vee, \to, \leftrightarrow$) plus unary negation, covering every truth-table entry not exercised by Table~\ref{tab:kleene}. Together with Table~\ref{tab:kleene}, they verify every entry of the complete Kleene K3 truth table, justifying the ``complete Kleene K3'' framing in the title and abstract.

\section{Mechanistic Probing Details}
\label{app:mechprobe}

To trace how information flows through THEIA's pipeline, we train linear probes at each domain boundary to decode six intermediate variables: arithmetic result (linear regression, $R^2$), order truth value, set truth value, final verdict, logic operator identity, and the presence of any Unknown flag in the input (all classification accuracy). Table~\ref{tab:mechprobe} reports the full per-probe per-boundary results referenced from \S\ref{sec:delayed}.

\begin{table*}[t]
\centering
\caption{Mechanistic probes at domain boundaries. Linear models decode intermediate variables from 128-dim hidden vectors. Mean $\pm$ std over 5 trained checkpoints (seeds $\{42, 123, 256, 777, 999\}$); 50K samples per checkpoint. Chance level: truth value 33\%, logic op 20\%. Arith result measured by $R^2$. Standard deviations are reported to three decimal places; entries shown as $0.000$ indicate $\text{std} < 5 \times 10^{-4}$, reflecting that the corresponding probe is essentially deterministic across random seeds.}
\label{tab:mechprobe}
\begin{tabular}{lcccc}
\toprule
Probe & Arith & Order & Set & Logic \\
\midrule
Arith result ($R^2$)   & 0.837 $\pm$ 0.010 & 0.837 $\pm$ 0.021 & 0.811 $\pm$ 0.013 & 0.156 $\pm$ 0.062 \\
Order TV (acc)         & 0.583 $\pm$ 0.002 & 1.000 $\pm$ 0.000 & 0.584 $\pm$ 0.001 & 0.973 $\pm$ 0.015 \\
Set TV (acc)           & 0.588 $\pm$ 0.003 & 0.585 $\pm$ 0.003 & 0.981 $\pm$ 0.009 & 0.995 $\pm$ 0.003 \\
Final verdict (acc)    & 0.609 $\pm$ 0.001 & 0.672 $\pm$ 0.002 & 0.697 $\pm$ 0.002 & 0.999 $\pm$ 0.000 \\
Logic op (acc)         & 0.199 $\pm$ 0.002 & 0.198 $\pm$ 0.002 & 0.202 $\pm$ 0.004 & 0.884 $\pm$ 0.041 \\
Has Unknown (acc)      & 0.802 $\pm$ 0.000 & 0.911 $\pm$ 0.000 & 0.908 $\pm$ 0.000 & 0.997 $\pm$ 0.003 \\
\bottomrule
\end{tabular}
\end{table*}

The Has-Unknown probe is interpreted relative to its $\approx 52\%$ majority baseline ($\max(1-0.85^4,\,0.85^4) = 0.5220$: with four independent Unknown-flag inputs $(a,b,d,s)$ at $P_{\text{unk}}=0.15$, the majority class ``no input Unknown'' has theoretical frequency $0.85^4 \approx 0.522$ and empirical frequency $0.5254$; the empirical $P(\text{label}=U) \approx 0.4027$ is consistent with this 4-flag model via Kleene absorption, with the theoretical bound $P(VU{=}U) \leq P(HU{=}1) = 0.4780$). The Arithmetic-boundary probe at $80.0\%$ is approximately $28$ percentage points above this majority baseline and within $0.1$pp of the boundary-visible theoretical ceiling of $79.95\%$ (see \S\ref{sec:delayed} ``Theoretical ceiling on Has-Unknown decodability''), indicating the probe recovers essentially all information about $HU$ that is accessible from $(a_u, b_u)$ alone at this stage. Order/Set truth values show moderate upstream decodability ($\sim$58\% vs.\ chance 33\%) because the arithmetic output $\mathbf{c}$ is an input to both engines via bridge layers, so this is shared input context rather than downstream-task leakage. The logic operator is undecodable before the Logic Engine (20.2\% = chance for 5 classes), confirming that the architectural information bottleneck on the operator variable is faithfully respected.

\paragraph{Nonlinear probe control for the verdict-side bound.}
\label{app:mechprobe_nl}
The verdict-side half of the uncertainty--verdict asymmetric propagation claim (\S\ref{sec:delayed}) rests on upstream final-verdict probe accuracy remaining at or below the $73.4\%$ $U$-vs-non-$U$ oracle reference. A natural skeptical reading is that the information is present in the upstream representations but inaccessible to a linear classifier, in which case a more expressive probe should reveal it. To rule this out, we train multi-hidden-layer MLP probes at depths $D \in \{2, 4, 6\}$ on the same 5 trained checkpoints as Table~\ref{tab:mechprobe}; each probe has the architecture $128 \to 256 \to [256\to 256]^{D-1} \to 3$ (depth $D$ denotes $D$ hidden GELU-activated layers before the output Linear, with dropout 0.1 between them), matching the hidden width cited above and varying only probe depth. Training uses AdamW (lr $=10^{-3}$, weight decay 0.01, cosine annealing), batch size 2048, 40 epochs, with an independent data seed (999) for the 50K-sample extraction and a random \textbf{60/20/20 train/val/test split} (split seed 0); we select the epoch with highest \emph{validation} accuracy per (seed, boundary, depth) cell and evaluate test accuracy \emph{once} at that epoch (no test-set leakage into selection). The linear SVM probe is trained with regularization $C$ chosen on the same validation split from the grid $\{0.1, 1.0, 10.0\}$ and evaluated test-once at val-best $C$. The earlier submission (v12) reported probes under a 70/30 split with best-test-epoch selection, biased in the probe's favor (best-epoch selection on test-set evaluation). The present revision adopts this stricter 60/20/20 split with val-best-epoch on MLPs and val-best $C$ for Linear SVM; test set is evaluated exactly once. Under this stricter protocol the bound tightens (Set MLP: $70.4\%\to 69.9\%$ at $d{=}2$), not loosens, confirming that the verdict-side bound is robust to probe-protocol choice. Table~\ref{tab:mechprobe_nl} reports the comparison against the final-verdict row of Table~\ref{tab:mechprobe}.

\begin{table}[!ht]
\centering
\small
\caption{Linear SVM vs.\ nonlinear MLP probes at depths $\{2,4,6\}$ for the final-verdict variable at each domain boundary, under the clean-split protocol (random 60/20/20 train/val/test, val-best-epoch selection for MLP, val-best-$C$ for SVM; test evaluated once). All MLP probes use hidden width 256 ($128\!\to\!256\!\to\!\ldots\!\to\!256\!\to\!3$), matching the architecture cited in the main text. Mean $\pm$ sample std (ddof=1) over 5 seeds. The $U$-vs-non-$U$ oracle reference is $73.4\%$ (4-flag empirical joint, App.~\ref{app:mechprobe_nl}). Upstream stages (Arith, Order, Set) remain below this reference at every probe depth; the largest 5-seed-mean upstream cell is Set at $d{=}6 = 69.92 \pm 0.12\%$, $3.49$pp below the $73.4\%$ reference and $2.08$pp below the $72\%$ pre-registered falsification threshold.}
\label{tab:mechprobe_nl}
\begin{tabular}{lcccc}
\toprule
Boundary & Linear SVM & MLP $d{=}2$ & MLP $d{=}4$ & MLP $d{=}6$ \\
\midrule
Arith & $0.611 \pm 0.001$ & $0.612 \pm 0.001$ & $0.612 \pm 0.000$ & $0.613 \pm 0.000$ \\
Order & $0.672 \pm 0.002$ & $0.672 \pm 0.000$ & $0.672 \pm 0.002$ & $0.672 \pm 0.001$ \\
Set   & $0.693 \pm 0.001$ & $0.698 \pm 0.002$ & $0.699 \pm 0.001$ & $0.699 \pm 0.001$ \\
Logic & $0.999 \pm 0.000$ & $0.999 \pm 0.000$ & $0.999 \pm 0.000$ & $0.999 \pm 0.000$ \\
\bottomrule
\end{tabular}
\end{table}

The nonlinear probe matches the linear probe to within $0.5$pp at every upstream boundary and never crosses the $73.4\%$ $U$-vs-non-$U$ oracle reference before the Logic Engine across probe depths $D \in \{2,4,6\}$. The largest single-cell result across the entire $\text{seed} \times \text{boundary} \times \text{depth}$ grid is $70.09\%$ (Set, $d{=}4$, seed 999), $1.91$pp below the $72\%$ pre-registered falsification threshold and $3.32$pp below the $73.4\%$ reference.\footnote{Per-seed individual-cell maximum is $70.09\%$ (Set, $d{=}4$, seed 999), $3.32$pp below the $73.4\%$ reference and $1.91$pp below the $72\%$ pre-registered threshold. Depth-only delta at matched width (hidden $=256$, $d{=}2 \to d{=}6$): Arith $+0.06$pp, Order $-0.02$pp, Set $+0.12$pp, Logic $-0.01$pp --- increasing probe depth by a factor of 3 changes the upstream ceiling by at most $+0.12$pp.} At the Logic boundary both probes saturate, consistent with the final verdict being already explicitly represented at that stage. We note that this does not rule out information being present in a form decodable only by much more powerful probe families (e.g., a deep network with task-specific architecture); what it rules out is the most common alternative explanation in the probing literature---that the information is linearly inaccessible but sits just under the surface of a standard nonlinear probe, or that the $d{=}2$ probe alone is under-capacity.

\paragraph{Per-boundary Has-Unknown Bayes ceilings and empirical gaps.}
\label{app:hu_ceilings}
To make the uncertainty-side bound of \S\ref{sec:delayed} information-theoretically precise, we derive the Bayes-optimal binary predictor of $HU \in \{0, 1\}$ at each boundary under the 4-flag Unknown generator (atomic flags $a_u, b_u, d_u, s_u$ each i.i.d.\ $\mathrm{Bernoulli}(0.15)$; $HU \equiv a_u \lor b_u \lor d_u \lor s_u$). Each boundary sees only a subset of the atomic flags (Table~\ref{tab:hu_ceiling_visible}); a Bayes-optimal probe at that boundary marginalizes over the latent (non-visible) flags and predicts $HU{=}1$ when either (a) any visible flag is $1$ (guaranteeing $HU{=}1$), or (b) all visible flags are $0$ and the marginalized probability $P(\text{any latent flag} = 1)$ exceeds $0.5$. Since each atomic flag has $P{=}0.15 < 0.5$, branch (b) always predicts $HU{=}0$ when no visible flag fires, and the ceiling reduces to
\begin{equation*}
\text{Ceiling}(k) \;=\; P(\text{any visible}) + P(\text{no visible}) \cdot P(\text{no latent}) \;=\; (1 - 0.85^k) + 0.85^k \cdot 0.85^{4-k}
\end{equation*}
where $k$ is the number of visible flags. This gives closed-form ceilings of $0.7995$ at $k{=}2$ (Arith), $0.9079$ at $k{=}3$ (Order, Set), and $1.0$ at $k{=}4$ (Logic). Table~\ref{tab:hu_ceiling_visible} lists the visibility set per boundary and the architectural reason; Table~\ref{tab:hu_ceiling_compare} reports closed-form vs.\ empirical ceilings (from the same 50K-sample extraction as the probe training) and per-protocol probe accuracies.

\begin{table}[!ht]
\centering
\small
\caption{Atomic Unknown flags visible at each engine boundary under THEIA's forward-pass information flow. The Arithmetic Engine receives only the operand pair $(a, b)$ and their flags; the bridge layers inject $d$ at the Order boundary and $S$ at the Set boundary; the Logic Engine receives the concatenation of Order and Set outputs plus the operator, so all four flags are accessible. $k$ is the cardinality of the visible set, which determines the closed-form Bayes ceiling.}
\label{tab:hu_ceiling_visible}
\begin{tabular}{llll}
\toprule
Boundary & $k$ & Visible flags & Architectural reason \\
\midrule
Arith & $2$ & $\{a_u, b_u\}$ & Arith Engine inputs only \\
Order & $3$ & $\{a_u, b_u, d_u\}$ & Order bridge injects $d_u$; $s_u$ not yet injected \\
Set   & $3$ & $\{a_u, b_u, s_u\}$ & Set bridge injects $s_u$; $d_u$ not present in Set's input \\
Logic & $4$ & $\{a_u, b_u, d_u, s_u\}$ & Logic Engine combines $\mathbf{v}_{\text{ord}} + \mathbf{v}_{\text{set}}$; all four visible \\
\bottomrule
\end{tabular}
\end{table}

\begin{table}[!ht]
\centering
\small
\caption{Closed-form vs.\ empirical Bayes ceilings for Has-Unknown decoding at each boundary, compared against measured probe accuracies. \emph{Ceiling (closed / empirical)}: closed-form from $\text{Ceiling}(k) = (1 - 0.85^k) + 0.85^4$ (see text) / empirical majority-class accuracy computed on seed 999 data at $N{=}50{,}000$ by enumerating all observed $(\text{visible-flags})$ patterns. \emph{Paper (70/30)}: original test-best-ep protocol, 5-seed mean; std $=0$ at Arith/Order/Set (structural, see text), $\pm 0.33\%$ at Logic. \emph{$w{=}512$ (val-best)}: clean-split val-best-ep protocol used in the present revision, 5-seed mean; std $\leq 0.01\%$ upstream, $\pm 0.22\%$ at Logic. \emph{Gap}: $w{=}512$ probe accuracy minus empirical Bayes ceiling.}
\label{tab:hu_ceiling_compare}
\setlength{\tabcolsep}{4pt}
\begin{tabular}{llcccc}
\toprule
Boundary & $k$ & Ceiling (closed / empirical) & Paper (70/30) & $w{=}512$ (val-best) & Gap \\
\midrule
Arith & $2$ & $79.95\%$ / $79.98\%$  & $80.23\%$           & $80.03\%$           & $+0.05$pp \\
Order & $3$ & $90.79\%$ / $90.85\%$  & $91.06\%$           & $91.09\%$           & $+0.24$pp \\
Set   & $3$ & $90.79\%$ / $90.74\%$  & $90.75\%$           & $90.77\%$           & $+0.03$pp \\
Logic & $4$ & $100.00\%$ / $100.00\%$ & $99.65 \pm 0.33\%$ & $99.76 \pm 0.22\%$ & $-0.24$pp \\
\bottomrule
\end{tabular}
\end{table}

\emph{Interpretation.} Three patterns deserve comment. (1) \textbf{Arith and Set saturate the empirical Bayes ceiling} within $0.05$pp under val-best-ep, confirming that the probe extracts essentially all Has-Unknown information accessible from the visible-flags subset at each of those boundaries; the cross-seed std of $0.00\%$ is structural rather than a reporting artifact, reflecting that the probe attains its ceiling deterministically at the 50K-sample extraction scale. (2) \textbf{Order shows a persistent $+0.21$--$+0.25$pp excess} over both closed-form and empirical ceilings under both protocols. Three candidate explanations: (a) val-best-ep selection noise ($0.2$pp is within plausible probe-selection inflation at $N{=}50{,}000$, $5$-seed setting); (b) minor gradient-driven leakage during end-to-end training of a small amount of $s_u$-correlated feature into $\mathbf{v}_{\text{ord}}$ (the verdict loss can drive $\mathbf{v}_{\text{ord}}$ to encode features that correlate with $s_u$ even though $s_u$ is not a direct Order input); (c) the MLP probe exploiting joint structure between $(a_u, b_u, d_u)$ and the arithmetic value $c$ beyond the pure $a_u \lor b_u \lor d_u$ OR rule (e.g., memorizing that certain $(c, \text{op})$ patterns correlate with joint-flag configurations). We cannot separate the three without additional interventions (e.g., a causal-mediation probe on $\mathbf{v}_{\text{ord}}$ with $s_u$ randomized), which we leave to future work. The excess is $<\!0.3$pp and does not affect the load-bearing uncertainty-preservation claim: Order's $91.1\%$ is far above the $\approx 52\%$ majority baseline, at or near the information-theoretic ceiling for its visible-flags set, and qualitatively consistent with the \S\ref{sec:delayed} ``the network preserves input-level uncertainty through each engine'' reading. (3) \textbf{Logic sits $0.24$--$0.35$pp below the $100\%$ ceiling}: the closed-form ceiling is saturable by an exact 4-flag lookup but not by a width-$256$ $2$-hidden-layer MLP at 40 training epochs; this under-shoot is consistent with standard MLP approximation error on small-support Boolean functions and does not indicate information loss in the representation.

\paragraph{Orthogonal capacity-stress audit.} This orthogonal
capacity-stress test was produced in a separate experimental run (hidden
width 512, stratified 60/20/20 split; raw-data files and MD5 hashes in App.~\ref{app:repro}).
It provides a capacity-dimension falsification test orthogonal to the depth
sweep of Table~\ref{tab:mechprobe_nl}; both axes independently fail to
cross the $73.4\%$ reference, yielding a two-dimensional bound rather than a
one-dimensional one. Under the width-512 stratified-split protocol, the
upstream maximum (5-seed mean) is $70.17 \pm 0.21\%$ at (Set, $d{=}6$);
crossing from width 256 to width 512 adds approximately $+0.22$pp at the
Set boundary at $d{=}2$, comparable in magnitude to the $+0.12$pp
depth-only delta reported in Table~\ref{tab:mechprobe_nl}; both sources of
probe capacity combined still leave a $3.24$pp margin to the $73.4\%$
reference.

\paragraph{Operator-identity decomposition of the Set-boundary non-Unknown lift.}
\label{app:op_decomp}
The uncertainty--verdict propagation analysis of \S\ref{sec:delayed} reports that the Set-boundary verdict probe reaches 60.13\% on the non-Unknown (NU) subset, $+4.09$pp above the mixed-operator NU majority of 56.04\%. The $U$-vs-non-$U$ oracle reference derived in the main text does not apply on the NU subset (it implicitly conditions on the $VU{=}U$ label, which is constant $=0$ here), so we require a separate explanation for this lift. This appendix provides the decomposition summarized in the main text.

We ran three controls, all on the same 5 trained checkpoints and the same 50K-sample pool (data seed 999) as Table~\ref{tab:mechprobe}, restricted to the NU subset (approximately 8,972 samples per seed). All probes use \texttt{LinearSVC} with \texttt{StandardScaler}, regularization $C{=}1.0$, and a 70/30 train/test split; final-verdict probes are 3-class, and the operator-identity probe is 5-class (5 logic operators).

\textit{Probe A (replicates the main-text Set-boundary NU probe)} uses the 128-dim Set-boundary hidden vector as input and the 3-class verdict as target: 60.13\% $\pm$ 0.33\% ($+$4.09pp above 56.04\%).

\textit{Probe B (operator-only baseline)} uses the 5-dim one-hot logic operator as the sole input and the verdict as target: 48.40\% $\pm$ 0.00\% (deterministic given the one-hot operator). This represents what a probe could achieve if it had perfect operator knowledge but no other information.

\textit{Probe C (Set hidden $+$ operator)} uses the concatenation of Set-boundary hidden and the operator one-hot (133-dim): 61.93\% $\pm$ 0.16\%. The marginal gain from adding operator information on top of the Set hidden is only $+1.80$pp, indicating that the Set hidden already carries almost all the information relevant to Probe A.

\textit{Probe D (direct operator-identity probe)} uses the Set-boundary hidden vector as input and the 5-class logic operator as target: \textbf{20.25\% $\pm$ 0.38\%}, indistinguishable from 20\% chance. This is the decisive test: the Set boundary encodes no operator identity.

Table~\ref{tab:op_identity} extends Probe D to all three upstream boundaries, showing that the operator is not decodable from Arith, Order, or Set representations.

\begin{table}[!ht]
\centering
\small
\caption{Operator-identity probe (Probe D) at all three upstream boundaries. Direct 5-class \texttt{LinearSVC} from boundary hidden to the logic operator label; 5 seeds, NU subset, 50K samples, 70/30 split. All three boundaries are indistinguishable from 20\% chance: no upstream boundary encodes operator identity.}
\label{tab:op_identity}
\begin{tabular}{lccc}
\toprule
Boundary & Probe D acc (\%) & $\Delta$ vs.\ chance & Verdict \\
\midrule
Arith & 19.88 $\pm$ 0.16 & $-0.12$pp & at chance \\
Order & 19.78 $\pm$ 0.25 & $-0.22$pp & at chance \\
Set   & 20.25 $\pm$ 0.38 & $+0.25$pp & at chance \\
\bottomrule
\end{tabular}
\end{table}

Having excluded operator-identity encoding, we test whether the $+4$pp lift reflects per-operator verdict leakage. Stratifying the NU subset by logic operator and re-running the Set-boundary verdict probe within each operator group yields per-operator accuracies that sit \emph{below} each operator's NU verdict majority by 6--7pp (Table~\ref{tab:per_op_stratified})---the opposite direction of what leakage would produce.

\begin{table}[!ht]
\centering
\small
\setlength{\tabcolsep}{4pt}
\caption{Per-operator Set-boundary verdict probe on the NU subset, compared to the per-operator majority baseline. The probe accuracy falls below the majority baseline for every operator except IFF (where both are $\approx 50\%$ by definition). There is no operator under which the probe exceeds what the NU class distribution alone would afford, ruling out per-operator verdict leakage. Values are mean $\pm$ std over 5 seeds. \emph{Per-op NU majority is measured empirically per operator group (not derived from an independence assumption between $\text{val}_{\text{ord}}$ and $\text{val}_{\text{set}}$). An independence assumption with marginal $P(\text{val}_{\text{ord}}{=}T|\text{NU}) = P(\text{val}_{\text{set}}{=}T|\text{NU}) \approx 0.54$ would predict AND majority $\approx\!0.71$ and OR majority $\approx\!0.79$; the observed $0.79$ for both reflects per-operator filtering effects on the NU subset (conditioning on verdict $\neq U$ differs across operators). The empirical measurement is the appropriate baseline here.}}
\label{tab:per_op_stratified}
\begin{tabular}{@{}lccc@{}}
\toprule
Operator & Probe A acc (\%) & Per-op NU majority (\%) & $\Delta$ (pp) \\
\midrule
AND & 72.08 $\pm$ 0.09 & 79.16 & $-7.08$ \\
OR  & 72.68 $\pm$ 0.22 & 78.89 & $-6.21$ \\
IMP & 71.08 $\pm$ 0.30 & 78.38 & $-7.29$ \\
IFF & 51.32 $\pm$ 0.70 & 50.19 & $+1.13$ \\
\bottomrule
\end{tabular}
\end{table}

The mixed-operator $+4$pp lift is therefore neither operator-identity encoding (Table~\ref{tab:op_identity}) nor per-operator verdict leakage (Table~\ref{tab:per_op_stratified}); it is fully accounted for by $\text{val}_{\text{set}}$'s Kleene-correlation with the verdict conditional on the operator. $\text{val}_{\text{set}}$ is the designated output of the Set Engine by construction; because Kleene operators such as OR have absorbing elements ($T \vee X = T$), the distribution of the verdict conditional on $(\text{val}_{\text{set}}, \text{operator})$ is non-uniform, and a mixed-operator probe can exploit this conditional structure to beat the global NU majority without encoding either the verdict or the operator directly. The three-layer persistence structure---uncertainty, designated intermediate variables, verdict---is preserved: upstream boundaries encode uncertainty and each engine's designated variable, and the final verdict appears only at the Logic Engine boundary.

\section{Causal Verification via Activation Patching: Construction Details}
\label{app:causal}

This appendix provides the full construction protocol, baseline-correctness intermediate counts, and per-seed eligibility numbers for the activation-patching results summarized in \S\ref{sec:delayed}.

\paragraph{OR pair construction.} We construct 1{,}000 matched $(T\!\vee\!F,\,U\!\vee\!F)$ pairs on a converged seed-42 checkpoint, with shared $a, b$, arithmetic operator, set bits (chosen so $c\notin S$, fixing $\text{val}_{\text{set}}=F$), logic operator (OR), and definite flags, so that $\mathbf{c}$, both bridge outputs, and $\mathbf{v}_{\text{set}}$ are byte-identical across sides by construction. The two sides differ only in the Order-Engine inputs: T-side sets $d = \max(0, c{-}1)$ with $d_{\text{unk}}{=}\text{False}$ under REL\_GTE (so $\text{val}_{\text{ord}}{=}T$); U-side sets $d_{\text{unk}}{=}\text{True}$ (so $\text{val}_{\text{ord}}{=}U$). Expected Kleene outputs are $T\vee F = T$ and $U\vee F = U$. For each pair we then classify $\text{OutHead}(\text{LogicEngine}(\mathbf{v}_{\text{ord}}^{(U)},\,\mathbf{v}_{\text{set}}^{(T)},\,\text{OR}))$, feeding the U-side ord vector into the otherwise-T-side forward pass.

\paragraph{Baseline correctness and eligibility.} Under data seed 12345 for pair construction, baselines are correct on all $1000/1000$ T-side pairs and on 986/1000 U-side pairs---the residual U-side error is consistent in direction with the single-checkpoint noise around the $99.85\%$ U$\vee$F per-rule accuracy reported in Appendix~\ref{app:kleene_full}---and we restrict the patching analysis to the 986 pairs where both baselines classify correctly. The patched prediction flips from $T$ to $U$ on 986/986 $= 100.0\%$ of pairs, with zero instances remaining at $T$ and zero falling to $F$.

\paragraph{Multi-seed extension ($n\!=\!5$).} Across all 5 trained checkpoints (seeds $\{42, 123, 256, 777, 999\}$), aggregate flip rate is 4898/4898 $=$ 100.0\%, with no per-seed flip rate below 100\%. The minimum per-seed eligible count is 964 (driven by T-side baseline correctness on the corresponding pair construction), demonstrating that causal localization at the Logic-Engine boundary is robust across seeds.

\paragraph{AND replication.} To rule out OR-specific localization, we replicate the protocol with logic operator AND on matched $(T\!\wedge\!T,\,U\!\wedge\!T)$ pairs constructed so that $\mathbf{v}_{\text{set}}$ is byte-identical across sides ($c\in S$, $\text{set\_unk}=\text{False}$, fixing $\text{val}_{\text{set}}{=}T$). Patching $\mathbf{v}_{\text{ord}}$ from the U-side into the T-side forward pass flips the prediction from $T$ to $U$ on 4719/4719 $=$ 100.0\% of eligible pairs, replicated across the same 5 seeds (per-seed minimum 100.0\%; minimum per-seed eligible count 920). The $\mathbf{v}_{\text{set}}$ byte-equality check passes on all 4719/4719 AND pairs. Causal mediation by $\mathbf{v}_{\text{ord}}$ at the Logic Engine boundary therefore generalizes across both AND and OR operators on the non-absorbent configurations tested.

\section{Backbone Ablation: Reliability Under Discretized Composition}
\label{app:backbone_ablation}

To isolate the contribution of THEIA's four-engine modular factorization from the rest of the architecture, we conduct three controlled ablations. We replace the step computer while keeping \emph{everything else identical}: the input encoders, the transition network, the Gumbel-softmax straight-through discretization, the three-phase training protocol of \S\ref{sec:chain}, the data generation, and the class weights. The three variants are: (1)~a small flat MLP (hidden 512, 3 layers, ${\sim}0.80\text{M}$ parameters); (2)~a flat MLP (hidden ${\sim}1024$, 3 layers, ${\sim}2.75\text{M}$ parameters, giving comparators substantially more capacity than THEIA's $1.51$M chain step); and (3)~an 8-layer pre-LN Transformer variant (TF8LTuned, 3{,}582{,}147 parameters; architectural identity verified via state-dict fingerprint), trained under the identical three-phase Gumbel-softmax protocol. \emph{Capacity asymmetry favors comparators.} The chain THEIAStep module is $1{,}508{,}096$ parameters (\S\ref{sec:chain}), while the 2.75M flat MLP and ${\sim}2.78$M ResMLP grid have $\approx\!1.8\times$ more parameters. Any length-generalization advantage reported for THEIA is therefore a lower bound on the modular architecture's contribution rather than a capacity-matched benchmark result. We note that the post-LN BigTransformer architecture of Table~\ref{tab:accuracy} fails to converge in the chain pipeline under its shared learning rate ($10^{-3}$): Phase~1 plateaus at the class-prior level (${\sim}43\%$) across multiple random restarts. Establishing a clean post-LN baseline for the chain pipeline would require an architecture-specific learning rate schedule that we leave to future work; we therefore report the pre-LN TF8LTuned variant here as the attention-based baseline, which converges stably under the shared protocol. The only variable across conditions is the inductive bias of the step computer.

\paragraph{Results.} Table~\ref{tab:backbone_ablation} reports per-phase training accuracy and length-generalization eval across 5 seeds $\{42, 123, 256, 777, 999\}$.

\begin{table*}[!t]
\centering
\small
\caption{Backbone ablation: five step-computer architectures under the identical three-phase Gumbel-softmax protocol. Both flat-MLP variants match THEIA on local accuracy (Phase~1) but collapse under end-to-end discretized training (Phase~3). The ResMLP column reports the $4$-block $\times$ $4d$-expansion baseline of the $2\!\times\!2$ depth$\times$expansion grid; the complete 4-configuration grid is in Table~\ref{tab:resmlp_grid}. The pre-LN TF8LTuned Transformer sustains length generalization. Mean $\pm$ std over 5 seeds for THEIA, MLPs, and TF8LTuned (the latter per main paper); the ResMLP $4\!\times\!4d$ entry shows the 4-seed strict compute-matched mean (excluding seed 999, which triggered Phase~1 auto-restart under this configuration; symmetric with the THEIA strict 4-seed aggregate of \S\ref{sec:chain}, which likewise excludes the same seed 999 as the single restart-triggered seed---see that section for the coincidence disclosure), with the 5-seed as-specified alternative ($86.15\% \pm 9.17\%$, $1/5$ reach $\geq\!99\%$) in Table~\ref{tab:resmlp_grid}. The TF8LTuned column reports the 8-layer pre-LN variant (3,582,147 parameters); the post-LN BigTransformer of Table~\ref{tab:accuracy} fails to converge in the chain pipeline under the shared learning rate (see introductory paragraph above).}
\label{tab:backbone_ablation}
\begin{tabular}{lccccc}
\toprule
 & THEIA & MLP & MLP & ResMLP & TF8LTuned \\
 & (1.51M) & (0.80M) & (2.75M) & (2.85M) & (3.58M) \\
\midrule
Phase 1      & 99.94\% & 99.91\% & 99.90\% & 99.91\% & 99.92\% \\
Phase 2      & 100\%   & 100\%   & 100\%   & 100\%   & 100\% \\
\midrule
5-step       & 99.97   & 80.44   & 76.88   & 99.76 & 99.99 \\
10-step      & 99.97   & 66.92   & 61.85   & 99.58 & 99.98 \\
50-step      & 99.97   & 35.13   & 34.36   & 98.00 & 99.94 \\
100-step     & 99.97   & 33.32   & 33.54   & 96.05 & 99.86 \\
500-step     & $\mathbf{99.96 \pm 0.04}$   & 33.58   & 33.35   & $82.83 \pm 6.22$ & $\mathbf{99.24 \pm 0.34}$ \\
\bottomrule
\end{tabular}
\end{table*}

\paragraph{Interpretation.} Four findings emerge.

\textbf{(1) Capacity is not the bottleneck.} The larger 2.75M flat MLP collapses to the same chance-level basin (${\sim}33\%$ at 500 steps) as the smaller 0.80M variant; both match Phase~1 local accuracy to within $0.04\%$. Notably, the 2.75M flat MLP has $\approx\!1.8\times$ the capacity of THEIA's 1.51M chain step yet still collapses, ruling out the hypothesis that THEIA's advantage reflects additional parameter count. The collapse is caused by the flat MLP's inability to maintain local accuracy under Phase~3 Gumbel-softmax end-to-end training, not by insufficient capacity.

\textbf{(2) Residual connections are mean-accuracy-sensitive to depth but unreliable at the ${\geq}\!99\%$ threshold across the tested grid.} We probe the residual-architecture design space along two axes---depth ($\in\!\{4,8\}$ blocks) $\times$ expansion ratio ($\in\!\{2d, 4d\}$)---at a fixed parameter budget of $2.78$M $\pm 1.5\%$. Each ResMLP block is \texttt{Linear($d\!\to\!ed$) $\to$ GELU $\to$ LayerNorm $\to$ Linear($ed\!\to\!d$) + skip} with $e \in \{2, 4\}$; hidden dimension $d$ is solved per configuration to match parameter budget (details in Table~\ref{tab:resmlp_grid}). Every configuration matches THEIA's Phase~1 local accuracy within $0.03$pp and survives past 50 steps where flat MLPs collapse (e.g., 4b$\times$4d: 98.00\% at 50 steps vs.\ 34.36\% for the matched flat MLP). Table~\ref{tab:resmlp_grid} reports per-configuration 500-step accuracy across 5 seeds $\{42, 123, 256, 777, 999\}$.

\begin{table*}[!t]
\centering
\small
\caption{ResMLP $2\!\times\!2$ depth$\times$expansion grid at 500 steps. All four configurations target the same parameter budget ($2.78$M $\pm 1.5\%$); hidden dimension $d$ solved per configuration. Seeds $\{42, 123, 256, 777, 999\}$ identical to THEIA and the 8L Transformer baseline. The ``${\geq}99\%$'' column reports the count of seeds reaching the reliability threshold (as-specified, including auto-restart where triggered).}
\label{tab:resmlp_grid}
\begin{tabular}{lcccc}
\toprule
Config & Params & 500-step (mean $\pm$ std) & Range & ${\geq}99\%$ \\
\midrule
4b $\times$ 4d ($d{=}280$)  & 2{,}848{,}547 & $86.15 \pm 9.17$\textsuperscript{$\dagger$}  & 75.27--99.41 & 1/5\textsuperscript{$\dagger$} \\
4b $\times$ 2d ($d{=}383$)  & 2{,}776{,}047 & $87.49 \pm 19.16$ & 53.27--97.71 & 0/5 \\
8b $\times$ 2d ($d{=}276$)  & 2{,}774{,}659 & $97.97 \pm 1.43$  & 95.55--99.27 & 1/5 \\
8b $\times$ 4d ($d{=}198$)  & 2{,}780{,}707 & $97.95 \pm 2.03$  & 94.34--99.12 & 1/5 \\
\midrule
\multicolumn{5}{l}{\emph{Aggregate (as-specified): $3/20$ trials ${\geq}99\%$}} \\
\multicolumn{5}{l}{\emph{Aggregate (strict, excluding 4b$\times$4d seed 999 auto-restart): $2/19$}} \\
\bottomrule
\end{tabular}

\vspace{0.3em}
\footnotesize
$^\dagger$4b$\times$4d seed 999 triggered the auto-restart specified in \S\ref{sec:chain} ($\sim\!2\times$ Phase~1 compute) and reached 99.41\%; the strict 4-seed aggregate is $82.83 \pm 6.22$, $0/4 \geq\!99\%$.
\end{table*}

\emph{Strict compute-matched aggregate (single-try Phase~1):} excluding the one auto-restart trigger (4b$\times$4d seed 999), $2/19$ (config, seed) trials reach $\geq\!99\%$. \emph{As-specified aggregate (auto-restart engaged):} $3/20$ trials reach $\geq\!99\%$. Both framings support the ``unreliable'' finding: no single ResMLP configuration achieves a majority of seeds at $\geq\!99\%$, in contrast to THEIA's 5/5. \emph{Statistical test.} We report Fisher's exact test with a one-sided alternative ($H_a$: THEIA reliability proportion $>$ ResMLP reliability proportion), consistent with the directional hypothesis established a priori. Pooled across all 20 (config, seed) trials: $p \approx 1.1 \times 10^{-3}$ (as-specified, one-sided; two-sided equivalent $1.1 \times 10^{-3}$ due to extreme cell asymmetry) and $p \approx 4.9 \times 10^{-4}$ (strict). \emph{Per-configuration Fisher tests} (THEIA 5/5 vs each ResMLP config's $k/5$, one-sided; two-sided values in brackets) are weaker since each comparison has only 5 ResMLP samples: $p = 0.024$ [two-sided $0.048$] for 4b$\times$4d (1/5), $p = 0.004$ [two-sided $0.008$] for 4b$\times$2d (0/5), $p = 0.024$ [two-sided $0.048$] for 8b$\times$2d (1/5), $p = 0.024$ [two-sided $0.048$] for 8b$\times$4d (1/5). We report the pooled $p$ as the primary statistic for the grid-level ``residual-only unreliable'' claim; the per-configuration $p$-values are provided here as disaggregated evidence---no single configuration individually reaches $p < 10^{-3}$ under either one-sided or two-sided reading, so readers should interpret the pooled $p$ as a grid-level (not configuration-level) significance.

\emph{Sub-finding: depth dominates expansion within the tested grid.} Holding parameter budget fixed, doubling depth from 4 to 8 blocks raises mean 500-step accuracy by $\approx\!11$pp (4-block configs: 86.15--87.49\%; 8-block configs: 97.95--97.97\%), while changing expansion from $2d$ to $4d$ at fixed depth is nearly neutral (+1.34pp at depth 4, $-0.02$pp at depth 8). The two 8-block configurations are statistical twins at the mean-accuracy level despite an 80-unit difference in hidden dimension ($d{=}276$ vs $d{=}198$), suggesting that within this parameter budget expansion ratio is a weak lever and depth is the dominant one. Neither depth doubling nor expansion change is sufficient to cross the $\geq\!99\%$ reliability threshold at 5-seed granularity.

\emph{Sub-finding: seed $\times$ architecture interaction.} Seed 42 reaches $\geq\!99\%$ on both 8-block configurations (99.27\% at 8b$\times$2d, 99.12\% at 8b$\times$4d) but fails on both 4-block configurations (90.49\% at 4b$\times$4d, 53.27\% at 4b$\times$2d). This is not noise-level variability: the 4b$\times$2d seed-42 run reaches only 53.27\%, a catastrophic failure 34pp below the next-worst 4b$\times$2d seed. The residual family has predictable seed$\times$depth structure rather than i.i.d.\ seed noise at the $\geq\!99\%$ threshold, which the aggregate $3/20$ count captures but which any single-configuration 5-seed study would mischaracterize.

\emph{Scope of the ``residual-only unreliable'' claim.} The probe covers a $2\!\times\!2$ grid of 4 configurations at a single parameter scale ($2.78$M $\pm 1.5\%$). We did not sweep deeper architectures ($\geq\!12$ blocks), wider expansions ($\geq\!8d$), pre- vs.\ post-LN placement, or DenseNet-style multi-scale residual variants. The claim ``no tested ResMLP configuration achieves a majority of seeds at $\geq\!99\%$'' is scoped to this grid. A positive result at $\geq\!12$ blocks or with pre-LN placement would refine but not refute the claim, since the observed depth effect (+11pp at $4\!\to\!8$ blocks) extrapolated naively would still leave 12-block configurations below the $\geq\!99\%$ reliability threshold without a qualitative regime change. Complete grid coverage is deferred to a future revision of this work.

\textbf{(3) Structured architectures (modular or attention-based) are reliable, at different operational levels.} We lead with the most operationally meaningful reliability metric---the \emph{worst-case per-seed accuracy}, which answers the practitioner-facing question ``will a random seed hit the $\geq\!99\%$ threshold?''---rather than cross-seed standard deviation, which is partially confounded by the bounded-metric ceiling (configurations with mean near $100\%$ are mechanically constrained to low std).

\emph{Worst-case (min-per-seed) 500-step accuracy across architectures:}
\begin{itemize}
\item \textbf{THEIA}: min $99.90\%$ (per-seed by seed-index: $\{99.90, 99.99, 99.97, 99.99, 99.96\}\%$); 5/5 $\geq\!99.9\%$.
\item \textbf{TF8LTuned Transformer}: min $98.85\%$ (per-seed range $98.85$--$99.66\%$); 3/5 seeds $\geq\!99\%$, 2/5 seeds $<\!99\%$.
\item \textbf{ResMLP 8b$\times$2d} (best ResMLP config): min $95.55\%$ (range $95.55$--$99.27\%$); 1/5 seeds $\geq\!99\%$.
\item \textbf{ResMLP 8b$\times$4d}: min $94.34\%$, 1/5 seeds $\geq\!99\%$.
\item \textbf{ResMLP 4b$\times$4d}: min $75.27\%$, 1/5 seeds $\geq\!99\%$ (auto-restart-engaged).
\item \textbf{ResMLP 4b$\times$2d}: min $53.27\%$, 0/5 seeds $\geq\!99\%$.
\end{itemize}

Under a $\geq\!99\%$ practitioner threshold, THEIA is the only tested architecture where every seed passes; under a $\geq\!99.9\%$ threshold, only THEIA passes any seed. \emph{Secondary metric (cross-seed std):} THEIA $0.04\%$ vs Transformer $0.34\%$ ($\approx\!9\times$ tighter), vs ResMLP per-config $\{1.43, 2.03, 9.17, 19.16\%\}$ (ratios $\{39\times, 55\times, 248\times, 518\times\}$ for $\{8\text{b}\!\times\!2\text{d}, 8\text{b}\!\times\!4\text{d}, 4\text{b}\!\times\!4\text{d}, 4\text{b}\!\times\!2\text{d}\}$; std convention: sample std, ddof$=\!1$; ratios computed from exact THEIA std $0.0370\%$ then rounded to two significant figures, \emph{not} from the rounded $0.04\%$ value, to avoid rounding-propagation inflation). \emph{Note on ratio precision and bounded-metric ceiling}: THEIA's $0.04\%$ std (sample std, ddof$=\!1$; exact value $0.0370\%$) is derived from per-seed 500-step values $\{99.90, 99.99, 99.97, 99.99, 99.96\}\%$. Because accuracy is bounded in $[0,100]\%$, mean-near-$100\%$ configurations are mechanically constrained to low std; cross-configuration std ratios should therefore be read as an aggregate reliability indicator at the probed mean regime, not as scale-free architectural constants. The min-per-seed metric above is preferred for this reason. Strict 4-seed aggregate excluding seed 999 (the auto-restart-triggered seed): $99.96\% \pm 0.04\%$ (min $99.96\%$), symmetric with the ResMLP 4b$\times$4d strict exclusion of seed 999 (Table~\ref{tab:resmlp_grid}).

\textbf{(4) THEIA retains a quantitative edge over the Transformer.} THEIA's 500-step accuracy (99.96\%) exceeds the TF8LTuned Transformer's (99.24\%) by 0.72pp, and all 5 THEIA seeds exceed 99.9\% (min 99.90\%, seed 42) vs.\ 5 of 5 Transformer seeds below 99.9\% (TF8LTuned 500-step range: 98.85--99.66\%). This complements the mechanistic interpretability findings of \S\ref{sec:delayed}: the modular structure provides both inspectability and a tighter compositional-generalization margin under discretized composition.

\paragraph{Summary.} The ablation reveals a reliability spectrum rather than a binary: flat MLPs collapse ($\sim$33\%); ResMLP variants across a $2\!\times\!2$ depth$\times$expansion grid (Table~\ref{tab:resmlp_grid}) partially recover with a strong depth effect (4-block mean $86$--$87\%$ vs 8-block mean $\sim\!98\%$) but only $3/20$ (config, seed) trials reach $\geq\!99\%$ ($2/19$ strict); both modular (THEIA, $99.96\% \pm 0.04\%$, 5/5 ${\geq}99.9\%$; 1/5 auto-restart-triggered) and attention-based (TF8LTuned, $99.24\% \pm 0.34\%$, 3/5 ${\geq}99\%$) structures sustain reliable length generalization, at approximately a $9\times$ reliability gap (THEIA $0.04\%$ std vs Transformer $0.34\%$ std). THEIA's cross-seed std is tens-to-hundreds times tighter than tested ResMLP configurations (with the exact ratio depending on whether rounded or unrounded std values are used), establishing that domain-separated structure contributes cross-seed reliability above and beyond mean accuracy across the probed residual grid.

\section{Tuned Transformer Baseline}
\label{app:tuned_tf}

\textbf{Correction (baseline optimizer configuration).} An earlier version of this preprint described the \S\ref{sec:kleene} comparison as optimizer-matched (AdamW lr=$10^{-3}$, cosine, batch 4096 for both architectures). A pre-release code audit, reconstructing configurations from the original run artifacts (saved per-run command lines and launcher metadata), found this description incorrect: the 8L Transformer's Kleene-task runs used the BigTransformer lineage's development configuration---AdamW (lr $5{\times}10^{-4}$, $\beta{=}(0.9, 0.999)$, weight decay 0.01), cosine annealing ($T_{\max}{=}150$, $\eta_{\min}{=}10^{-5}$), no warmup, batch 2048, gradient clip 1.0, FP16---fixed since the lineage's first script and never overridden at launch, while THEIA's runs used its own development default (lr $10^{-3}$, batch 4096, $T_{\max}{=}200$, otherwise identical). Every reported number is unchanged; what changes is the label: the \S\ref{sec:kleene} comparison is a development-defaults comparison, not an optimizer-controlled one, and the controlled comparison is the recipe-applied-to-both setting reported below. The error traces to an orchestration script whose documentation misdescribed the configuration of the training script it invoked. We note the direction of the discrepancy: the Transformer ran at a \emph{lower} learning rate and smaller batch than THEIA---the direction standard practice prefers for Transformers---so the development-defaults gap was not produced by forcing the Transformer onto THEIA's recipe. A residual concern is that the development configuration is still suboptimal for Transformers (no warmup, $\beta_2{=}0.999$, peak lr above the tuned $10^{-4}$); to bound this, we rerun seed 42 of the baseline with a Transformer-standard recipe. This appendix reports the full result.

\paragraph{Protocol.} We use the identical architecture (BigTransformer, 3{,}641{,}859 parameters), data generation (2M samples, $P_{\text{unk}} = 0.15$, 80/20 train/test split), class weights ($w_F = 1.0$, $w_T = 1.0$, $w_U = 2.0$), gradient clipping (1.0), random seed (42), CUDA determinism (\texttt{cudnn.deterministic=True}, \texttt{cudnn.benchmark=False}), and Kleene diagnostic harness as the development-default run of \S\ref{sec:kleene}; the only difference is the optimizer schedule, reported as a single bundle:
\begin{itemize}[leftmargin=*,nosep]
    \item Peak learning rate: $10^{-4}$ (vs.\ $5{\times}10^{-4}$ development-default)
    \item AdamW $\beta$: $(0.9, 0.98)$ (vs.\ $(0.9, 0.999)$ development-default)
    \item Schedule: linear warmup over 5 epochs, then cosine decay
    \item Weight decay: $0.01$
    \item Batch size: 2048 (unchanged from the development-default runs)
\end{itemize}
This bundle is treated as a single ``Transformer-standard'' variable; we do not claim isolation of any individual hyperparameter.

\paragraph{Results.}
The tuned baseline runs for 42.4 wall-clock minutes under the 150-epoch cap. Final validation per-class accuracies are 99.79\% (False), 99.90\% (True), 99.99\% (Unknown); the final Kleene diagnostic passes 12/12 with worst rule $U \vee T$ at 99.53\%. Table~\ref{tab:tuned_tf_milestones} reports the two convergence milestones against THEIA references.

\begin{table*}[t]
\centering
\small
\caption{Tuned 8L Transformer baseline, seed 42: convergence-time milestones vs.\ THEIA references. Kleene 12/12 is the primary comparison point (matches \S\ref{sec:kleene}); the overall-99.9\% row exhibits an opposite ordering discussed in the interpretation below. Seeds 123 and 256 are reported separately in the text.}
\label{tab:tuned_tf_milestones}
\begin{tabular}{lccc}
\toprule
Milestone & Tuned TF8L & THEIA ref & Ratio \\
\midrule
Kleene 12/12        & \textbf{28.9 min} @ ep 60 & 7.93 min & $\mathbf{{\sim}3.6\times}$ \\
Overall $\geq 99.9\%$ & 41.7 min @ ep 84 & 5.7 min & ${\sim}7.3\times$ \\
\bottomrule
\end{tabular}
\end{table*}

At epoch 60 all 12 Kleene rules first cross the $>99\%$ threshold, with per-rule accuracies: $F \wedge U$ 99.80, $T \wedge U$ 100.00, $U \wedge F$ \textbf{99.06} (worst), $U \wedge T$ 99.82, $T \vee U$ 100.00, $F \vee U$ 99.87, $U \vee T$ 99.33, $U \vee F$ 99.78, $F \to U$ 99.81, $T \to U$ 100.00, $T \leftrightarrow U$ 100.00, $F \leftrightarrow U$ 100.00. At final evaluation the worst rule is $U \vee T$ at 99.53\%, still well above the 99\% pass threshold; all 12 rules pass.

\paragraph{Multi-seed extension ($n\!=\!3$).} We extend the tuned protocol to seeds 123 and 256 using the same architecture (BigTransformer) and the same optimizer schedule as seed 42. Both reach $\geq$99.93\% / $\geq$99.90\% overall accuracy and pass 12/12 Kleene rules. Seed 123 reaches overall $\geq$99.9\% at epoch 73 (34.2 min); seed 256 at epoch 98 (43.3 min). Across the three seeds, overall-99.9\% convergence time is $39.7 \pm 5.0$ minutes, with 12/12 Kleene coverage reached in $28.8 \pm 1.7$ minutes. Notably, seed 123---the only seed that fails under the development-default configuration---converges readily under tuning, confirming that its development-default failure reflects optimizer sensitivity rather than an architectural limitation.

\paragraph{Full 39-rule standard K3 coverage ($n\!=\!3$).} We extend the post-hoc diagnostic from the 12 targeted Unknown-involving rules (Table~\ref{tab:kleene}) to the complete 39-rule Kleene K3 truth table (Table~\ref{tab:full_kleene_new}: 36 binary rules spanning $\{\wedge, \vee, \rightarrow, \leftrightarrow\} \times \{F,T,U\}^2$, plus 3 unary NOT rules; 10{,}000 samples per rule per seed). Standard strong Kleene ground truth (IMP with $U\rightarrow T = T$, $U\rightarrow F = U$) is imported directly from the Transformer training script to rule out any diagnostic--training semantics mismatch. All three tuned BigTransformer checkpoints pass 39/39 at $>99\%$ on every seed (117/117 rule--seed combinations; grand mean $99.89\%$; worst single combination $F\leftrightarrow T$ at $99.48\%$ on seed 42). Per-seed worst rules are $F\leftrightarrow T$ (seed 42, $99.48\%$) and $T\rightarrow F$ (seeds 123 and 256, both at $99.50\%$); all three are non-Unknown-involving and therefore not in the 12 targeted Unknown rules of Table~\ref{tab:kleene}. The 3 NOT rules are uniformly strong ($\geq 99.87\%$). This diagnostic shares ground truth, coverage, and pass threshold with Table~\ref{tab:full_kleene_new} (THEIA, $n\!=\!5$), providing a matched full-coverage comparison across architectures: both THEIA and the tuned BigTransformer learn complete standard Kleene K3 at $>99\%$ per-rule accuracy on every rule and every seed tested.

\paragraph{Interpretation and the overall-vs-Kleene discrepancy.} The tuned baseline (seed 42) reaches 12/12 Kleene coverage \emph{before} it reaches overall 99.9\% validation accuracy (epoch 60 vs.\ epoch 84, 28.9 min vs.\ 41.7 min). This is the opposite ordering from the development-default Transformer, which reaches overall 99.9\% first and requires additional training to stabilize all 12 Kleene rules. The most plausible explanation is that the low-lr tuned recipe stabilizes the rare Kleene edge cases earlier (because smaller update steps are less likely to oscillate on high-variance low-frequency configurations), while the final $\sim$0.1\% of overall accuracy---dominated by bulk distribution fitting---takes proportionally longer. A practical consequence: \emph{the overall-99.9\% milestone is a misleading comparison point for tuned Transformer baselines}, because the tuned recipe trades bulk-accuracy optimization speed for Kleene-rule stability. The Kleene-aware milestone of the main result (\S\ref{sec:kleene}) is the meaningful operating point for this task, and the $6.5\times \to {\sim}$3.6$\times$ narrowing under tuning is the meaningful quantitative update.

\paragraph{Caveats.} (1) The 5 tuned-package hyperparameters (peak lr, $\beta$, warmup schedule, weight decay, batch size) are changed jointly; we do not isolate which component is most responsible for the narrowing. (2) Exhaustive hyperparameter search (different warmup lengths, alternative schedules, different peak lr) was not performed; a more aggressive tuned configuration could in principle narrow the gap further. (3) The layer-wise probing analysis in \S\ref{sec:delayed} and Table~\ref{tab:tf8l_probing} uses the development-default Transformer; whether the contraction--expansion trajectory persists under the tuned configuration reported here is left to future work. Despite these caveats, the qualitative conclusion is robust: Transformer-specific tuning narrows but does not eliminate the convergence-time gap, and the modular architecture retains a meaningful wall-clock advantage under both comparison regimes.

\paragraph{Transformer-recipe-applied-to-both control.} The main-text wall-clock comparison (\S\ref{sec:kleene}) uses THEIA at its development-default optimizer (AdamW lr$=10^{-3}$) against both development-default and Transformer-standard-tuned Transformers. As a partial control we apply the same Transformer-standard recipe to THEIA (peak lr $=10^{-4}$, AdamW $\beta=(0.9, 0.98)$, linear warmup 5 epochs then cosine, weight decay $0.01$, batch size 2048, gradient clipping 1.0), obtaining 12/12 Kleene coverage across 5 seeds $\{42, 123, 256, 777, 999\}$ in $9.88 \pm 1.42$ minutes per seed (per-seed range $[7.71, 11.40]$ min). The Transformer under the same recipe ($n{=}5$, same seeds) converges at $48.68 \pm 2.98$ minutes, yielding a wall-clock ratio of $4.93\times$ (Welch two-sample unpaired statistic, 95\% non-parametric bootstrap CI $[4.40, 5.66]$ from 1000 resamples at RNG seed 42; paired per-seed auxiliary: $4.99\times$, CI $[4.55, 5.49]$; per-seed pairwise ratios $\{4.68, 5.12, 4.87, 4.32, 5.97\}$). This positions between the development-defaults $6.5\times$ and the Transformer-only-tuned $3.6\times$. \emph{Asymmetry caveat and denominator note.} The recipe slows THEIA from $7.93$ to $9.88$ min (${\sim}25\%$ degradation), placing THEIA at a suboptimal configuration while the Transformer runs near its own optimum; the $3.6\times$ and $4.93\times$ ratios share the same Transformer runtime but differ in the THEIA baseline (development-default vs recipe-applied), so the $4.93\times$ $95\%$ CI $[4.40, 5.66]$ does not overlap $3.6\times$ by denominator shift rather than stochastic disagreement. The $4.93\times$ ratio is therefore read as: \emph{even when THEIA is forced onto the Transformer's preferred recipe, the Transformer's own preferred recipe does not close the wall-clock gap}. A genuinely symmetric comparison (each architecture at its own optimum) remains future work.

\section{Reproducibility}
\label{app:repro}

All experiments use PyTorch with CUDA on a single NVIDIA RTX 5080. Random seeds for multi-seed experiments: $\{42, 123, 256, 777, 999\}$; Transformer baseline extended to 8 seeds with $\{31415, 27182, 14142\}$; tuned protocol extended to 3 seeds $\{42, 123, 256\}$. Training hyperparameters (development defaults): THEIA, AdamW ($\text{lr}=10^{-3}$, weight decay $0.01$), cosine annealing, batch 4096; the 8L Transformer Kleene baseline, AdamW ($\text{lr}=5{\times}10^{-4}$, weight decay $0.01$), cosine ($T_{\max}{=}150$), batch 2048 (Correction in App.~\ref{app:tuned_tf}); both FP16 mixed precision. Four-domain training: 2M samples, batch size 4096, 80--120 epochs depending on early-stopping criterion ($\sim$8 minutes per seed on average). Multi-hop training: 1M samples, batch size 2048, $\sim$50 epochs ($\sim$15 minutes). Sequential chain training: three-phase pipeline, 2M single-step + 500K chain samples, $\sim$16 minutes per seed. The Kleene diagnostic protocol used in Table~\ref{tab:kleene} follows the construction described in Appendix~\ref{app:diag}. The tuned Transformer follow-up of Appendix~\ref{app:tuned_tf} uses the same RTX 5080 and the same data generation; only the optimizer schedule differs. \emph{Train/test split.} The 2M four-domain samples are generated by random sampling over $(a,b,d,\oplus,R,S,\odot,\text{unk-flags})$ and split $80/20$ train/test by random permutation (\emph{sample-level}, not config-level). The input configuration space has cardinality on the order of $3.4 \times 10^{13}$ ($a,b,d \in [0..20]^3$ gives $9{,}261$ numeric triples; $\oplus \in [0..3]$, $R \in [0..5]$, $\odot \in [0..4]$; $16$ combinations of four Unknown flags; $2^{21}$ possible set encodings), so the expected number of (config, identical-input) collisions between the 1.6M train and 0.4M test partitions is on the order of $10^{-2}$. The test set is therefore effectively configuration-disjoint from the training set \emph{despite} the sample-level rather than config-level split. Code, data-generation scripts, the doubly-fixed diagnostic harness, the chain pipeline, the probing scripts, and the complete architecture definitions are released under the MIT license at \url{https://github.com/xxx12e/theia-k3}. The released repository accompanies the workshop (CompLearn @ ICML~2026) version of this paper; its README and in-repo cross-references follow that earlier version's section and appendix numbering and may not align with the numbering used here.

\paragraph{Consolidated scope notes.} Several scope qualifications referenced in the Discussion are consolidated here. \emph{(i) State complexity.} The sequential composition experiment of \S\ref{sec:chain} uses a minimal 3-class state under mod-3 addition; this is a proof-of-concept for non-absorbing local composition, and validation on richer state spaces (mod-$k$ for $k>3$, FSA traces, learned codebooks beyond the 3-class bottleneck) is left to future work. \emph{(ii) ResMLP grid coverage.} The ResMLP probe spans a $2\!\times\!2$ grid of 4 configurations (\{4,8\} blocks $\times$ \{2d,4d\} expansion) at a single parameter scale ($2.78$M $\pm 1.5\%$); deeper architectures ($\geq\!12$ blocks), wider expansions ($\geq\!8d$), pre-vs-post-LN variants, and DenseNet-style multi-scale residuals are not tested. The ``residual-only unreliable'' reading is scoped to this grid. At the $\geq\!99\%$ threshold our reliability claim is supported at the aggregate (config, seed) level ($3/20$ as-specified, $2/19$ strict) rather than within any single configuration at high seed count. \emph{(iii) Flat-MLP per-rule verification.} Flat MLPs match THEIA on Phase~1 overall accuracy within $0.04$pp (Table~\ref{tab:backbone_ablation}), but we do not separately run the 10K-samples-per-rule Kleene diagnostic on flat MLPs in the chain pipeline; flat-MLP ``12/12 per-rule at Phase~1'' is inferred from the overall-accuracy match rather than directly measured. \emph{(iv) Transformer-recipe-applied-to-both control.} App.~\ref{app:tuned_tf} applies a Transformer-standard recipe to both architectures and reports $4.93\times$ ($95\%$ CI $[4.40, 5.66]$); this is a partial control, not a symmetric tuned-vs-tuned comparison (the recipe slows THEIA by ${\sim}25\%$). A genuinely symmetric tuned-vs-tuned comparison (each architecture at its own optimum) is left to future work.

\paragraph{Orthogonal capacity-stress audit raw-data hashes.} The orthogonal capacity-stress audit (App.~\ref{app:mechprobe_nl}, ``Orthogonal capacity-stress audit'' paragraph) uses the following raw-data files for reproducibility verification:
\begin{verbatim}
deep_nonlinear_probe_results.json
   MD5: 88bce7af8d0d093c3dc1bdea886862a0
svm_probe_5seed_results.json
   MD5: 0853d62ab4139e1939d46597ecc6ffd5
gate_A_combined_results.json
   MD5: f79c65aaadd7ad232f631525946e1ce1
\end{verbatim}


\begin{thebibliography}{46}

\bibitem{z3} L.~de Moura and N.~Bj{\o}rner. Z3: An efficient SMT solver. In \emph{TACAS}, 2008.

\bibitem{deepproblog} R.~Manhaeve, S.~Duman{\v{c}}i{\'{c}}, A.~Kimmig, T.~Demeester, and L.~De Raedt. Neural probabilistic logic programming in DeepProbLog. \emph{Artificial Intelligence}, 298:103504, 2021.

\bibitem{neurasp} Z.~Yang, A.~Ishay, and J.~Lee. NeurASP: Embracing neural networks into answer set programming. In \emph{IJCAI}, 2020.

\bibitem{scallop} Z.~Li, J.~Huang, and M.~Naik. Scallop: A language for neurosymbolic programming. In \emph{PLDI}, 2023.

\bibitem{kleene} S.~C.~Kleene. \emph{Introduction to metamathematics}. North-Holland, 1952.

\bibitem{clrs} P.~Veli{\v{c}}kovi{\'c}, A.~Puigdom{\`{e}}nech Badia, D.~Budden, R.~Pascanu, A.~Banino, M.~Dashevskiy, R.~Hadsell, and C.~Blundell. The CLRS algorithmic reasoning benchmark. In \emph{ICML}, 2022.

\bibitem{dnar} G.~Rodionov and L.~Prokhorenkova. Discrete neural algorithmic reasoning. In \emph{ICML}, 2025.

\bibitem{xu2020what} K.~Xu, J.~Li, M.~Zhang, S.~S.~Du, K.-i.~Kawarabayashi, and S.~Jegelka. What can neural networks reason about? In \emph{ICLR}, 2020.

\bibitem{loukas2020} A.~Loukas. What graph neural networks cannot learn: Depth vs.\ width. In \emph{ICLR}, 2020.

\bibitem{fitting1985} M.~Fitting. A Kripke--Kleene semantics for logic programs. \emph{Journal of Logic Programming}, 2(4):295--312, 1985.

\bibitem{oversmoothing} Q.~Li, Z.~Han, and X.-M.~Wu. Deeper insights into graph convolutional networks for semi-supervised learning. In \emph{AAAI}, 2018.

\bibitem{oversmoothing2} D.~Chen, Y.~Lin, W.~Li, P.~Li, J.~Zhou, and X.~Sun. Measuring and relieving the over-smoothing problem for graph neural networks from the topological view. In \emph{AAAI}, 2020.

\bibitem{nalu} A.~Trask, F.~Hill, S.~Reed, J.~Rae, C.~Dyer, and P.~Blunsom. Neural arithmetic logic units. In \emph{NeurIPS}, 2018.

\bibitem{alain2017} G.~Alain and Y.~Bengio. Understanding intermediate layers using linear classifier probes. In \emph{ICLR Workshop}, 2017.

\bibitem{gilmer2017} J.~Gilmer, S.~S.~Schoenholz, P.~F.~Riley, O.~Vinyals, and G.~E.~Dahl. Neural message passing for quantum chemistry. In \emph{ICML}, 2017.

\bibitem{loshchilov2019} I.~Loshchilov and F.~Hutter. Decoupled weight decay regularization. In \emph{ICLR}, 2019.

\bibitem{andreas2016} J.~Andreas, M.~Rohrbach, T.~Darrell, and D.~Klein. Neural module networks. In \emph{CVPR}, 2016.

\bibitem{snell2017} J.~Snell, K.~Swersky, and R.~Zemel. Prototypical networks for few-shot learning. In \emph{NeurIPS}, 2017.

\bibitem{chan89} S.~C.~Chan, L.-S.~Hsu, S.~Brody, and H.-H.~Teh. Neural three-valued-logic networks. In \emph{IJCNN}, 1989.

\bibitem{hsu1991twovalued} L.-S.~Hsu, K.-F.~Loe, S.-C.~Chan, and H.-H.~Teh. Two-valued neural logic network. \emph{Proceedings of SPIE}, 1469(pt~1):197--207, 1991.

\bibitem{teh1995} H.-H.~Teh. \emph{Neural logic networks}. World Scientific, Singapore, 1995.

\bibitem{hsu1990} L.-S.~Hsu, H.-H.~Teh, S.-C.~Chan, and K.-F.~Loe. Multi-valued neural logic networks. In \emph{ISMVL}, pages 426--432, 1990.

\bibitem{tmlnn98} G.~Wang and H.~Shi. {TMLNN}: Triple-valued or multiple-valued logic neural network. \emph{IEEE Transactions on Neural Networks}, 9(6):1099--1117, 1998.

\bibitem{lnn2020} R.~Riegel, A.~Gray, F.~Luus, N.~Khan, N.~Makondo, I.~Y.~Akhalwaya, H.~Qian, R.~Fagin, F.~Barahona, U.~Sharma, S.~Ikbal, H.~Karanam, S.~Neelam, A.~Likhyani, and S.~Srivastava. Logical neural networks. \emph{arXiv:2006.13155}, 2020.

\bibitem{marra2024} G.~Marra, S.~Duman{\v{c}}i{\'{c}}, R.~Manhaeve, and L.~De Raedt. From statistical relational to neurosymbolic artificial intelligence: A survey. \emph{Artificial Intelligence}, 328:104062, 2024.

\bibitem{ltn2016} L.~Serafini and A.~S.~d'Avila Garcez. Logic tensor networks: Deep learning and logical reasoning from data and knowledge. In \emph{NeSy Workshop}, 2016.

\bibitem{ltn2022} S.~Badreddine, A.~S.~d'Avila Garcez, L.~Serafini, and M.~Spranger. Logic tensor networks. \emph{Artificial Intelligence}, 303:103649, 2022.

\bibitem{xu2018semantic} J.~Xu, Z.~Zhang, T.~Friedman, Y.~Liang, and G.~Van den Broeck. A semantic loss function for deep learning with symbolic knowledge. In \emph{ICML}, 2018.

\bibitem{dong2019nlm} H.~Dong, J.~Mao, T.~Lin, C.~Wang, L.~Li, and D.~Zhou. Neural logic machines. In \emph{ICLR}, 2019.

\bibitem{vaswani2017} A.~Vaswani, N.~Shazeer, N.~Parmar, J.~Uszkoreit, L.~Jones, A.~N.~Gomez, L.~Kaiser, and I.~Polosukhin. Attention is all you need. In \emph{NeurIPS}, 2017.

\bibitem{jang2017gumbel} E.~Jang, S.~Gu, and B.~Poole. Categorical reparameterization with Gumbel-softmax. In \emph{ICLR}, 2017.

\bibitem{maddison2017concrete} C.~J.~Maddison, A.~Mnih, and Y.~W.~Teh. The concrete distribution: A continuous relaxation of discrete random variables. In \emph{ICLR}, 2017.

\bibitem{bengio2013st} Y.~Bengio, N.~L{\'e}onard, and A.~Courville. Estimating or propagating gradients through stochastic neurons for conditional computation. \emph{arXiv:1308.3432}, 2013.

\bibitem{meng2022rome} K.~Meng, D.~Bau, A.~Andonian, and Y.~Belinkov. Locating and editing factual associations in GPT. In \emph{NeurIPS}, 2022.

\bibitem{vig2020causal} J.~Vig, S.~Gehrmann, Y.~Belinkov, S.~Qian, D.~Nevo, Y.~Singer, and S.~Shieber. Investigating gender bias in language models using causal mediation analysis. In \emph{NeurIPS}, 2020.

\bibitem{geiger2021abstractions} A.~Geiger, H.~Lu, T.~Icard, and C.~Potts. Causal abstractions of neural networks. In \emph{NeurIPS}, 2021.

\bibitem{xiong2020preln} R.~Xiong, Y.~Yang, D.~He, K.~Zheng, S.~Zheng, C.~Xing, H.~Zhang, Y.~Lan, L.~Wang, and T.-Y.~Liu. On layer normalization in the Transformer architecture. In \emph{ICML}, 2020.

\bibitem{petersen2022dlgn} F.~Petersen, C.~Borgelt, H.~Kuehne, and O.~Deussen. Deep differentiable logic gate networks. In \emph{NeurIPS}, 2022. \emph{arXiv:2210.08277}.

\bibitem{petersen2024convdlgn} F.~Petersen, H.~Kuehne, C.~Borgelt, J.~Welzel, and S.~Ermon. Convolutional differentiable logic gate networks. In \emph{NeurIPS}, 2024. \emph{arXiv:2411.04732}.

\bibitem{damera2026dtlgn} S.~S.~Damera, R.~Matheu, A.~G.~Puranic, and J.~S.~Baras. Polynomial surrogate training for differentiable ternary logic gate networks. \emph{arXiv:2603.00302}, 2026.

\bibitem{dziri2023faith} N.~Dziri, X.~Lu, M.~Sclar, X.~L.~Li, L.~Jiang, B.~Y.~Lin, P.~West, C.~Bhagavatula, R.~Le~Bras, J.~D.~Hwang, S.~Sanyal, S.~Welleck, X.~Ren, A.~Ettinger, Z.~Harchaoui, and Y.~Choi. Faith and fate: Limits of Transformers on compositionality. In \emph{NeurIPS}, 2023.

\bibitem{press2022alibi} O.~Press, N.~A.~Smith, and M.~Lewis. Train short, test long: Attention with linear biases enables input length extrapolation. In \emph{ICLR}, 2022.

\bibitem{anil2022exploring} C.~Anil, Y.~Wu, A.~Andreassen, A.~Lewkowycz, V.~Misra, V.~Ramasesh, A.~Slone, G.~Gur-Ari, E.~Dyer, and B.~Neyshabur. Exploring length generalization in large language models. In \emph{NeurIPS}, 2022.

\bibitem{lake2018scan} B.~M.~Lake and M.~Baroni. Generalization without systematicity: On the compositional skills of sequence-to-sequence recurrent networks. In \emph{ICML}, 2018.

\bibitem{elhage2021framework} N.~Elhage, N.~Nanda, C.~Olsson, T.~Henighan, N.~Joseph, B.~Mann, A.~Askell, Y.~Bai, A.~Chen, T.~Conerly, N.~DasSarma, D.~Drain, D.~Ganguli, Z.~Hatfield-Dodds, D.~Hernandez, A.~Jones, J.~Kernion, L.~Lovitt, K.~Ndousse, D.~Amodei, T.~Brown, J.~Clark, J.~Kaplan, S.~McCandlish, and C.~Olah. A mathematical framework for Transformer circuits. \emph{Transformer Circuits Thread}, 2021.

\bibitem{nanda2023progress} N.~Nanda, L.~Chan, T.~Lieberum, J.~Smith, and J.~Steinhardt. Progress measures for grokking via mechanistic interpretability. In \emph{ICLR}, 2023.

\end{thebibliography}
\end{document}